\documentclass[lettersize,journal]{IEEEtran}
\usepackage{amsmath,amsfonts}
\usepackage{array}
\usepackage[caption=false,font=normalsize,labelfont=sf,textfont=sf]{subfig}
\usepackage{textcomp}
\usepackage{stfloats}
\usepackage{url}
\usepackage[ruled]{algorithm2e}
\usepackage{xcolor}
\usepackage{multirow}
\usepackage{verbatim}
\usepackage{graphicx}
\def\BibTeX{{\rm B\kern-.05em{\sc i\kern-.025em b}\kern-.08em
    T\kern-.1667em\lower.7ex\hbox{E}\kern-.125emX}}
\usepackage{balance}
\begin{document}
\title{Hierarchical Identity Learning for Unsupervised Visible-Infrared Person Re-Identification}
\author{IEEE Publication Technology Department}

\author{Haonan Shi,
    Yubin Wang,
    De Cheng,
    Lingfeng He,
	Nannan~Wang,~\IEEEmembership{Senior Member,~IEEE,}
    Xinbo~Gao,~\IEEEmembership{Fellow,~IEEE}

\thanks{
This work was supported in part by the National Natural Science Foundation of China under Grants 62176198, U22A2096, in part by the Key R\&D Program of Shaanxi Province under Grant 2024GX-YBXM135.}

\thanks{
Haonan Shi, De Cheng, Lingfeng He, Nannan Wang and Xinbo Gao are with the State Key Laboratory of Integrated Services Networks, School of Telecommunications Engineering, Xidian University, Xi’an 710071, Shanxi, P. R. China (email: dcheng@xidian.edu.cn, lfhe@stu.xidian.edu.cn, nnwang@xidian.edu.cn  and xbgao@mail.xidian.edu.cn).}

\thanks{
Yubin Wang is with the Department of Computer Science and Technology, Tongji University, Shanghai 201804, China (e-mail: wangyubin2018@tongji.edu.cn.} 

\thanks{
\emph{(Corresponding author: De Cheng).}}


}


\maketitle

\begin{abstract}
Unsupervised visible-infrared person re-identification (USVI-ReID) aims to learn modality-invariant image features from unlabeled cross-modal person datasets by reducing the modality gap while minimizing reliance on costly manual annotations. Existing methods typically address USVI-ReID using cluster-based contrastive learning, which represents a person by a single cluster center. However, they primarily focus on the commonality of images within each cluster while neglecting the finer-grained differences among them. To address the limitation, we propose a Hierarchical Identity Learning (HIL) framework. Since each cluster may contain several smaller sub-clusters that reflect fine-grained variations among images, we generate multiple memories for each existing coarse-grained cluster via a secondary clustering. Additionally, we propose Multi-Center Contrastive Learning (MCCL) to refine representations for enhancing intra-modal clustering and minimizing cross-modal discrepancies. To further improve cross-modal matching quality, we design a Bidirectional Reverse Selection Transmission (BRST) mechanism, which establishes reliable cross-modal correspondences by performing bidirectional matching of pseudo-labels. Extensive experiments conducted on the SYSU-MM01 and RegDB datasets demonstrate that the proposed method outperforms existing approaches. The source code is available at: https://github.com/haonanshi0125/HIL.
\end{abstract}


\section{Introduction}

\IEEEPARstart{V}{isible}-infrared person re-identification (VI-ReID)~\cite{feng2019learning,ye2020cross,Ye_2021_ICCV,OTLA,adca,zheng2022visible} is an important research direction in the field of computer vision, aiming to match the images of the same person between the visible and infrared modalities. Compared to the extensively researched single-modality person ReID~\cite{wang2022nformer,Cho_2022_CVPR}, VI-ReID is a more challenging task due to the large modality gap between visible and infrared images. However, annotating cross-modal datasets demands more resources than single-modal datasets. To tackle the challenge of heavy annotations on large-scale cross-modal data, several semi-supervised methods~\cite{OTLA,yang2023translation,Shi_2023_ICCV} have been proposed for visible-infrared person re-identification, leveraging both labeled and unlabeled data to learn modality-invariant and identity-discriminative representations. Although these methods have shown promising results, they still depend on a certain amount of annotated data. 

\begin{figure}[!t] 
    \centering 
    \includegraphics[width=0.97\linewidth]{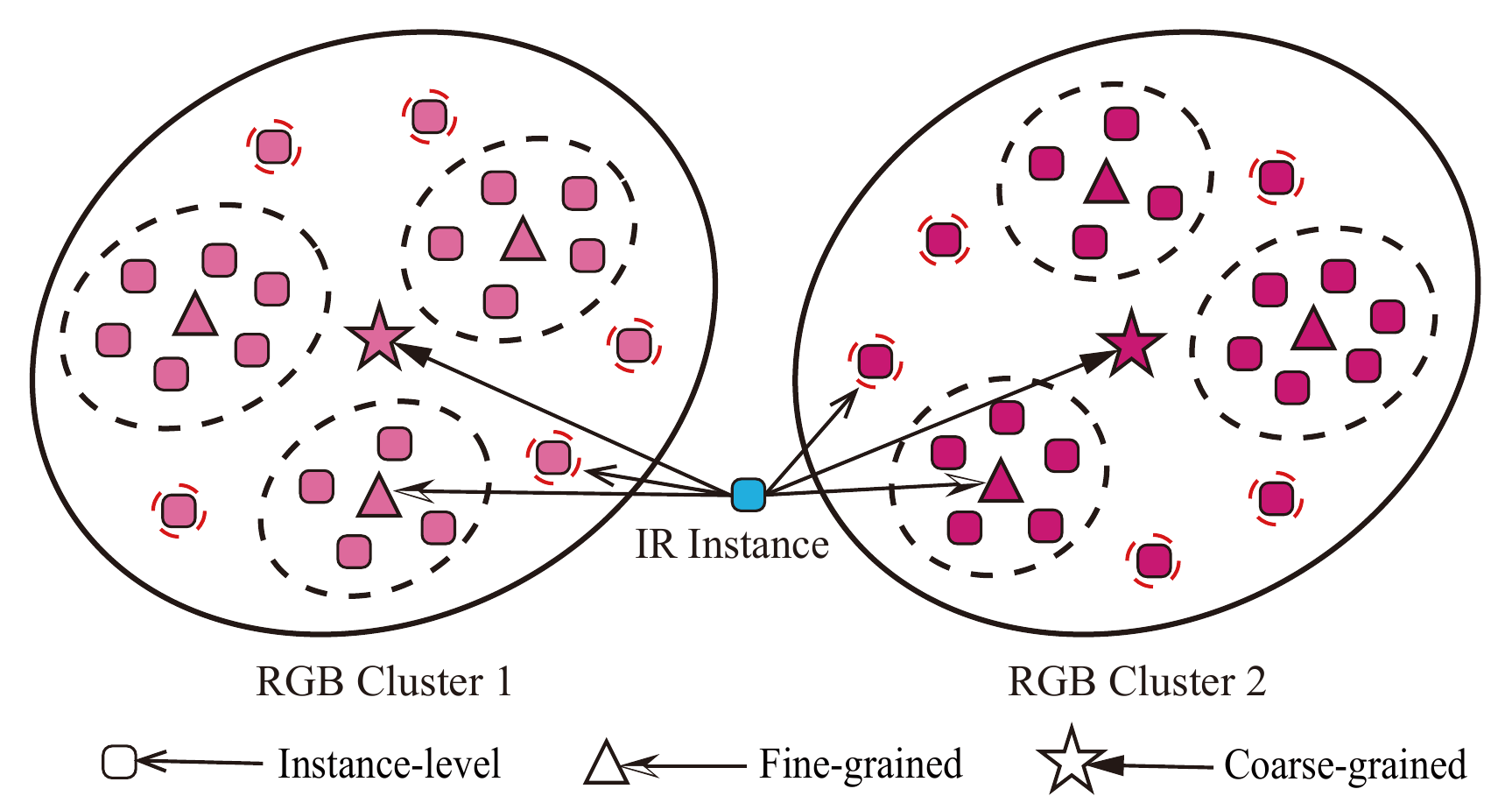} 
    \caption{Illustration of the hierarchical structure of feature alignment. Hierarchical identity information is extracted at three levels through two rounds of clustering, facilitating the refinement of representations within and across modalities via aligning instances with other instances, fine-grained centers, and coarse-grained centers.} 
    \label{fig:1} 
\end{figure}

To address scenarios lacking annotated data, some unsupervised methods~\cite{liang2021homogeneous,CHCR,MBCCM,he2024exploring,li2024romo} have been proposed, which employ a single memory bank for each cluster in both visible and infrared modalities. During training, the memory centers act as prototypes, and contrastive loss between query images and these prototypes, which are aggregated based on their similarity, is minimized. While single-memory representation learning offers simplicity, it may fail to capture the diverse perspectives and intricate details of the same identity, potentially resulting in information loss. To overcome this limitation, MMM~\cite{shi2025multi} employs multi-memory representations for identity generation, which enhances the capacity to capture fine-grained variations. However, simply relying on multi-memory representations without preserving coarse-grained information may lead to overfitting, as it neglects the macro-level commonality and informative features of the identity. As a result, the model tends to concentrate on a few specific fine-grained patterns, which can adversely affect optimization for modality-invariant features. As shown in Figure~\ref{fig:1}, we argue that incorporating a hierarchical structure of features, including instance-level, coarse-grained, and fine-grained features, enhances the refinement of representations and promotes better alignment both within and across modalities.

To this end, we propose a Hierarchical Identity Learning (HIL) framework, which collaboratively integrates both coarse-grained and fine-grained pseudo-labels. Specifically, our framework leverages a two-stage clustering strategy to refine the representation of identity features. In the first stage, we employ DBSCAN~\cite{ester1996density} to generate coarse-grained pseudo-labels, grouping visible and infrared instance features into identity clusters. However, as each cluster may contain multiple smaller sub-clusters that reflect fine-grained variations among images, a single clustering stage may fail to capture the intricate details and diverse perspectives of the same identity. To overcome this limitation, we introduce a secondary clustering step within each coarse-grained cluster, which facilitates a more comprehensive understanding of identity features.

To effectively utilize and model the hierarchical identity information, we propose a Multi-Center Contrastive Learning (MCCL) strategy. By leveraging the centers of multi-granularity clusters as reference points, MCCL constructs robust positive and negative sample sets. This approach allows the model to refine representations through contrastive learning both within and across modalities, thereby enhancing intra-modal clustering and minimizing cross-modal discrepancies. Furthermore, we design a Bidirectional Reverse Selection Transmission (BRST) mechanism to improve cross-modal matching quality. This mechanism performs bidirectional matching of pseudo-labels between visible and infrared modalities, enabling the establishment of reliable cross-modal correspondences. A reverse selection process is integrated to filter out unreliable pseudo-label matches, thereby enhancing their overall quality and robustness.

By combining these components, our proposed HIL framework addresses key limitations and achieves improved performance in cross-modal retrieval tasks. Our contributions can be summarized as follows:
\begin{itemize}
    \item We propose a Hierarchical Identity Learning (HIL) framework that employs a secondary clustering step within coarse-grained clusters to capture fine-grained identity variations, enabling robust representation.
    \item We propose a Multi-Center Contrastive Learning (MCCL) strategy that utilizes hierarchical identity information to construct robust positive and negative sample sets and refine representations for enhancing intra-modal clustering and minimizing cross-modal discrepancies.
    \item We design a Bidirectional Reverse Selection Transmission (BRST) mechanism that performs bidirectional matching of pseudo-labels between modalities, ensuring robust cross-modal correspondences by filtering unreliable matches through a reverse selection process.
    \item Extensive experiments conducted on the SYSU-MM01 and RegDB datasets demonstrate that the proposed method outperforms existing approaches in various settings.
\end{itemize}

\section{Related Work}

\subsection{Supervised Visible-Infrared Person ReID}
Supervised visible-infrared person ReID methods primarily focus on bridging the gap between the two different modalities using labeled person images. MPANet~\cite{wu2021discover} proposed a joint modality and pattern alignment network to discover cross-modality nuances. CAJ~\cite{ye2021channel} proposed a channel augmented joint learning strategy to improve the robustness against color variations by randomly exchanging the color channels. CIFT~\cite{li2022counterfactual} proposed a counterfactual intervention feature transfer method to address the balance gap between training-test modality and suboptimal topology structure problems. FMCNet~\cite{zhang2022fmcnet} employed GANs to compensate for missing modality-specific information at the feature level. TransVI~\cite{chai2023dual} designed a transformer-based visible-infrared network with a two-stream structure to capture modality-specific features and learn shared knowledge. SEFL~\cite{Feng_2023_CVPR} proposed a shape-erased feature learning paradigm to eliminate body-shape-related information from the learned features. SAAI~\cite{Fang_2023_ICCV} proposed a semantic alignment and affinity inference framework to explore the joint application of semantic-aligned feature learning and the affinity inference method. IDKL~\cite{ren2024implicit} proposed an implicit discriminative knowledge learning network to uncover and leverage the implicit discriminative information contained within the modality-specific. These methods have effectively reduced the modality gap, demonstrating their usefulness for supervised VI-ReID. However, the performance of these methods requires extensive human-labeled cross-modal data, which is time-consuming and expensive.

\subsection{Unsupervised Single-Modality Person ReID}
Unsupervised single-modality person ReID tasks endeavor to learn robust representations for unlabeled person images within a single modality. In order to mitigate the effects of noisy pseudo-labels, MMT~\cite{ge2020mutual} introduced a mutual mean-teaching framework that provides reliable soft pseudo-labels. SPCL~\cite{ge2020self} gradually generates more robust clusters through a self-paced contrastive learning framework with hybrid memory. ICE~\cite{Chen_2021_ICCV} introduced an inter-instance contrastive encoding method that enhances cluster compactness and improves pseudo-labels quality. RLCC~\cite{zhang2021refining} proposed a method to accurately estimate pseudo-label similarities between consecutive training generations using clustering consensus, and to refine pseudo-labels through temporal propagation and ensembling. IICS~\cite{xuan2021intra} proposed a two-stage similarity computation strategy that separately models intra-camera and inter-camera relations to generate more reliable pseudo-labels. To further improve the quality of the pseudo-labels, ISE~\cite{zhang2022implicit} designed an implicit sample extension method to strengthen the reliability of clusters. PPLR~\cite{Cho_2022_CVPR} proposed a part-based pseudo-label refinement framework that reduces pseudo-label noise by utilizing the complementary relationship between global and part features. Cluster-Contrast~\cite{Dai_2022_ACCV} proposed a cluster contrast method that stores unique centroid features and performs contrastive learning at the cluster level. However, they are difficult to directly apply to solving unsupervised visible-infrared person ReID tasks due to the large modality gap.

\subsection{Unsupervised Visible-Infrared Person ReID}
Unsupervised visible-infrared person ReID tasks aim to learn modality-invariant representations and establish reliable cross-modal associations between visible and infrared modalities without identity annotations. ADCA~\cite{adca} proposed an augmented dual-contrastive aggregation learning framework based on the ideas of homogeneous joint learning and heterogeneous aggregation. CHCR~\cite{CHCR} introduced a cross-modality hierarchical clustering and refinement method by enhancing modality-invariant feature learning and strengthening the reliability of pseudo-labels. DOTLA~\cite{cheng2023unsupervised} employed the optimal transport strategy to assign pseudo-labels from one modality to another modality at the instance level. PGM~\cite{PGM} proposed a progressive graph matching method to establish reliable cross-modal correspondences. SDCL~\cite{yang2024shallow} developed a shallow-deep collaborative learning framework based on the transformer architecture, which reduces the modality gap through the collaboration of shallow and deep features. PCLHD~\cite{shi2024learning} introduced a method of progressive contrastive learning with hard and dynamic prototypes, effectively learning commonality, divergence, and variety. MMM~\cite{shi2025multi} proposed a multi-memory matching framework to effectively capture intra-class nuances and establish reliable cross-modality correspondences. RPNR~\cite{yin2024robust} introduced a robust pseudo-label learning with neighbor relation framework, enhancing pseudo-label reliability and strengthening cross-modality alignment. Although these approaches demonstrate satisfactory performance, they are insufficient for effectively learning modality-invariant features without the simultaneous integration of multi-level identity information.

\section{Method}

\begin{figure*}[ht] 
    \centering 
    \includegraphics[width=\linewidth]{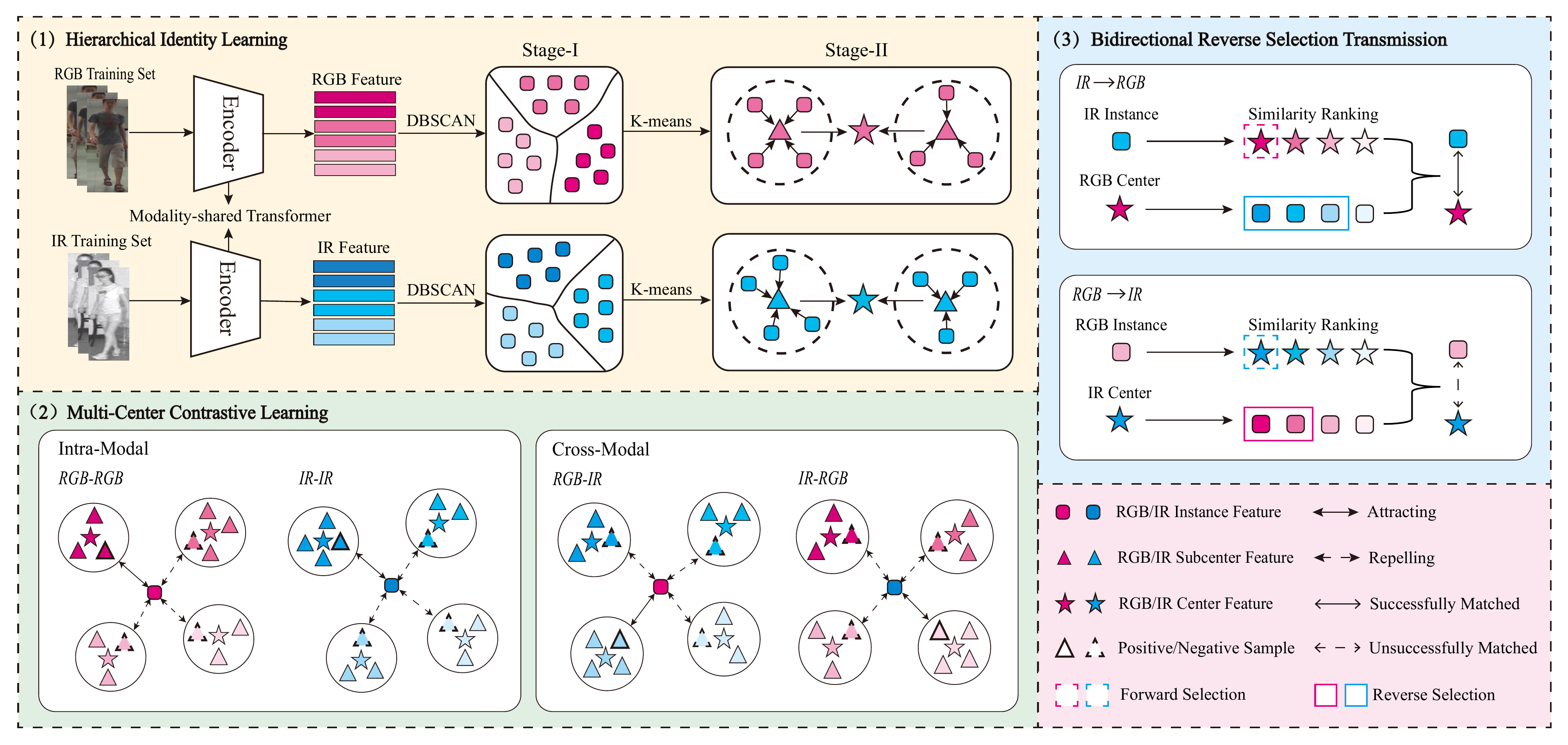} 
    \caption{Illustration of the Hierarchical Identity Learning (HIL) framework. Given unlabeled visible-infrared data, this framework generates more comprehensive hierarchical identity information, with the Multi-Center Contrastive Learning (MCCL) strategy utilizing these centers to construct robust positive and negative sample sets for refining representations via contrastive learning. For the cross-modal alignment, the Bidirectional Reverse Selection Transmission (BRST) mechanism performs bidirectional matching of pseudo-labels between two modalities.} 
    \label{fig:2} 
\end{figure*}

\subsection{Problem Formulation and Overview}

Given an unlabeled visible-infrared ReID dataset $\mathcal{X} = \left\{ {\mathcal{X}}^v, {\mathcal{X}}^r \right\}$, where ${\mathcal{X}}^v = \left\{ \mathbf{x}^v_k \right\}_{k=1}^{N_v}$ denotes the visible dataset containing $N_v$ unlabeled samples and ${\mathcal{X}}^r = \left\{ \mathbf{x}^r_k \right\}_{k=1}^{N_r}$ denotes the infrared dataset containing $N_r$ unlabeled samples. Our objective is to train a deep neural network $f_{\theta}(\cdot)$ that maps images from both modalities into a shared feature space. This shared feature space is designed to produce $d$-dimensional representations such that $\mathbf{f}_k^v = f_{\theta}(\mathbf{x}_k^v) \in \mathbb{R}^d$ and $\mathbf{f}_k^r = f_{\theta}(\mathbf{x}_k^r) \in \mathbb{R}^d$. Following the mapping process, we can obtain the instance feature set $\mathcal{F} = \left\{ {\mathcal{F}}^v, {\mathcal{F}}^r \right\}$, where ${\mathcal{F}}^v = \left\{ \mathbf{f}^v_k \right\}_{k=1}^{N_v}$ denotes the $N_v$ visible instance features and ${\mathcal{F}}^r = \left\{ \mathbf{f}^r_k \right\}_{k=1}^{N_r}$ denotes the  $N_r$ infrared instance features. By ensuring the representations are modality-invariant, the network enables effective matching of images of the same identity across the two modalities.

Our proposed Hierarchical Identity Learning (HIL) framework, illustrated in Figure \ref{fig:2}, integrates the Multi-Center Contrastive Learning (MCCL) strategy and the Bidirectional Reverse Selection Transmission (BRST) mechanism to enhance identity representation learning. This framework is built upon a dual-path transformer architecture inspired by ~\cite{yang2024shallow}. Within this architecture, instance features are clustered in two stages to produce coarse-grained and fine-grained pseudo-labels. Based on the identity information, MCCL refines representations via contrastive learning on multiple selected centers, aiming to derive robust and reliable intra-modal representations and minimize the modality gap. Additionally, BRST refines cross-modal matching by establishing consistent cross-modal correspondences through bidirectional pseudo-label matching in a reverse manner. By integrating hierarchical identity learning with the cross-modal feature matching algorithm, HIL effectively addresses the challenge of feature robustness and identity alignment across modalities. 

\subsection{Hierarchical Identity Learning}

\paragraph{Coarse-grained Pseudo-Label Generation} To obtain pseudo-labels in the unsupervised setting, we employ DBSCAN~\cite{ester1996density} for clustering visible and infrared instance features into $M_v$ and $M_r$ groups, respectively, at the first stage. After clustering, we can obtain the coarse-grained pseudo-label set $\mathcal{Y}^c = \left\{ {\mathcal{Y}}^{vc}, {\mathcal{Y}}^{rc} \right\}$, where ${\mathcal{Y}}^{vc} = \left\{\mathbf{y}^{vc}_i\right\}_{i=1}^{M_v}$ represents the $M_v$ visible coarse-grained pseudo-labels and ${\mathcal{Y}}^{rc} = \left\{\mathbf{y}^{rc}_i\right\}_{i=1}^{M_r}$ represents the $M_r$ infrared coarse-grained pseudo-labels. By calculating the mean feature of all instance features within each corresponding coarse-grained cluster, we can obtain a set of coarse-grained cluster features $\mathcal{U}^c = \left\{\mathcal{U}^{vc}, \mathcal{U}^{rc}\right\}$, where ${\mathcal{U}}^{vc} = \left\{\mathbf{u}^{vc}_i\right\}_{i=1}^{M_v}$ represents the $M_v$ visible coarse-grained cluster features and ${\mathcal{U}}^{rc} = \left\{\mathbf{u}^{rc}_i\right\}_{i=1}^{M_r}$ represents the $M_r$ infrared coarse-grained cluster features. The coarse-grained cluster features are calculated as follows:

\begin{equation}
\mathbf{u}^{vc}_i = \frac{1}{|\mathcal{H}^{vc}_i|} \sum_{\mathbf{f}^{vc}_n \in \mathcal{H}^{vc}_i} \mathbf{f}^{vc}_n,
\end{equation}
\begin{equation}
\mathbf{u}^{rc}_i = \frac{1}{|\mathcal{H}^{rc}_i|} \sum_{\mathbf{f}^{rc}_n \in \mathcal{H}^{rc}_i} \mathbf{f}^{rc}_n,
\end{equation}

\noindent where $\mathbf{f}^{vc}_n$ and $\mathbf{f}^{rc}_n$ are visible and infrared instance features within the same coarse-grained pseudo-label, respectively. $\mathcal{H}^{vc}_i$ and $\mathcal{H}^{rc}_i$ are visible and infrared coarse-grained cluster sets, respectively. The operator $|\cdot|$ counts the number of samples of a set.

During each training iteration, the coarse-grained cluster features of both modalities are updated using a momentum-based strategy:

\begin{equation}
\mathbf{u}^{vc}_{i,t} \leftarrow \alpha \mathbf{u}^{vc}_{i,t-1} + (1 - \alpha) q^v, \quad q^v \in \mathcal{H}^{vc}_i,
\end{equation}

\begin{equation}
\mathbf{u}^{rc}_{i,t} \leftarrow \alpha \mathbf{u}^{rc}_{i,t-1} + (1 - \alpha) q^r, \quad q^r \in \mathcal{H}^{rc}_i,
\end{equation}

\noindent where \( q^v \) and \( q^r \) are visible and infrared query features. The \( \alpha \) is a momentum hyperparameter. $t$ and $t-1$ refer to the current and last iterations, respectively.

Given visible and infrared query features $q^{v}$ and $q^{r}$, we compute the contrastive loss for visible and infrared modalities by the following equations:

\begin{equation}
\mathcal{L}_{id}^{v} = -\log \frac{\exp \left( q^{v} \cdot \mathbf{u}_{+}^{vc} / \tau \right)}{\sum_{i=0}^{M_v} \exp \left( q^{v} \cdot \mathbf{u}_{i}^{vc} / \tau \right)},
\end{equation}

\begin{equation}
\mathcal{L}_{id}^{r} = -\log \frac{\exp \left( q^{r} \cdot \mathbf{u}_{+}^{rc} / \tau \right)}{\sum_{i=0}^{M_r} \exp \left( q^{r} \cdot \mathbf{u}_{i}^{rc} / \tau \right)},
\end{equation}

\noindent where $\mathbf{u}_{+}^{vc}$ and $\mathbf{u}_{+}^{rc}$ are the positive coarse-grained cluster features corresponding to the pseudo-labels of $q^v$ and $q^r$, respectively. The $\tau$ is a temperature hyperparameter. The overall identity loss is defined as:

\begin{equation}
\mathcal{L}_{id} = \mathcal{L}_{id}^{v} + \mathcal{L}_{id}^{r}.
\end{equation}

\paragraph{Fine-grained Pseudo-Label Generation} After the first stage, the clustering process may not capture the diverse perspectives and intricate details associated with the same identity. To further explore fine-grained variations within existing identity clusters, we apply a secondary clustering algorithm, such as K-means~\cite{macqueen1967some}. This secondary clustering step is performed within each coarse-grained cluster to refine the identity representation. As a result, we can obtain a fine-grained pseudo-label set $\mathcal{Y}^f = \left\{ \mathcal{Y}^{vf}, \mathcal{Y}^{rf}\right\}$, where $\mathcal{Y}^{vf} = \left\{\mathbf{y}^{vf}_{ij} \right\}_{i=1,j=1}^{i=M_v,j=K}$ represents the visible fine-grained pseudo-label set and $\mathcal{Y}^{rf} = \left\{\mathbf{y}^{rf}_{ij}\right\}_{i=1,j=1}^{i=M_r,j=K}$ represents the infrared fine-grained pseudo-label set. Specifically, $K$ is the number of fine-grained clusters within each coarse-grained cluster. By calculating the mean feature of all instance features within each corresponding fine-grained cluster, we can obtain a fine-grained cluster feature set $\mathcal{U}^f = \left\{\mathcal{U}^{vf}, \mathcal{U}^{rf}\right\}$, where $\mathcal{U}^{vf} = \left\{\mathbf{u}^{vf}_{ij}\right\}_{i=1,j=1}^{i=M_v,j=K}$ represents the visible fine-grained cluster features and $\mathcal{U}^{rf} = \left\{\mathbf{u}^{rf}_{ij} \right\}_{i=1,j=1}^{i=M_r,j=K}$ represents the infrared fine-grained cluster features. The fine-grained cluster features are calculated as follows:

\begin{equation}
\mathbf{u}^{vf}_{ij} = \frac{1}{|\mathcal{H}^{vf}_{ij}|} \sum_{\mathbf{f}^{vf}_n \in \mathcal{H}^{vf}_{ij}} \mathbf{f}^{vf}_n,
\end{equation}
\begin{equation}
\mathbf{u}^{rf}_{ij} = \frac{1}{|\mathcal{H}^{rf}_{ij}|} \sum_{\mathbf{f}^{rf}_n \in \mathcal{H}^{rf}_{ij}} \mathbf{f}^{rf}_n,
\end{equation}

\noindent where $\mathbf{f}^{vf}_n$ and $\mathbf{f}^{rf}_n$ are visible and infrared instance features within the same fine-grained pseudo-label, respectively. $\mathcal{H}^{vf}_{ij}$ and $\mathcal{H}^{rf}_{ij}$ are visible and infrared fine-grained cluster sets, respectively. The operator $|\cdot|$ counts the number of samples in a set.

The hierarchical identity learning framework enables a comprehensive analysis of identities and facilitates the convergence of instance features toward their corresponding explicit centers. This process enhances the formation of positive and negative pairs, which are critical for the subsequent multi-center contrastive learning strategy.

\subsection{Neighbor Contrastive Learning}

Based on the obtained hierarchical identity information, we further expand the contrastive learning by exploring the relationship between different instances. For each visible instance feature, we calculate its cosine similarity with all visible instance features to find the one with the highest similarity. The cosine similarity between the instance feature $\mathbf{f}_k^{v}$ and the instance feature $\mathbf{f}_n^{v}$ is computed as follows:
\begin{equation}
\text{Sim}(\mathbf{f}_k^{v}, \mathbf{f}_n^{v}) = \frac{\mathbf{f}_k^{v} \cdot \mathbf{f}_n^{v}}{\|\mathbf{f}_k^{v}\|\|\mathbf{f}_n^{v}\|}.
\end{equation}

The instance feature with the highest similarity for a given instance feature $\mathbf{f}_k^v$ can be identified as:
\begin{equation}
\tilde{\mathbf{f}}^v_k = \underset{\mathbf{f}_n^{v} \in \mathcal{F}^{v}}{\arg\max}\, \text{Sim}(\mathbf{f}_k^v, \mathbf{f}_n^{v}).
\end{equation}

Next, we select reliable neighbors as the positive sample set $\mathcal{R}_k^v$ and choose the remaining visible instance features as the negative sample set $\mathcal{W}_k^v$ for the instance feature $\mathbf{f}_k^v$. The $\mathcal{R}_k^v$ and $\mathcal{W}_k^v$ can be obtained by:

\begin{equation}
\mathcal{R}_k^v= \left\{ \mathbf{f}_n^v \,\middle|\, Sim(\mathbf{f}_{k}^v, \mathbf{f}_n^v) > \beta \cdot Sim(\mathbf{f}_{k}^v, \tilde{\mathbf{f}}^v_k) \right\},
\end{equation}

\begin{equation}
\mathcal{W}_k^v= \left\{ \mathbf{f}_n^v \,\middle|\, Sim(\mathbf{f}_{k}^v, \mathbf{f}_n^v) \leq \beta \cdot Sim(\mathbf{f}_{k}^v, \tilde{\mathbf{f}}^v_k) \right\},
\end{equation}

\noindent where $\beta$ is a selection threshold hyperparameter. 

By obtaining the positive and negative sample sets, the contrastive loss for visible-visible $\mathcal{L}^{vv}_{neighbor}$ is defined as follows:

\begin{align}
S^v_{k,r} &= \sum_{\mathbf{f} \in \mathcal{R}_k^v} \exp\left(\frac{\text{Sim}(\mathbf{f}_k^v, \mathbf{f})}{\tau}\right), \\
S^v_{k,w} &= \sum_{\mathbf{f} \in \mathcal{W}_k^v} \exp\left(\frac{\text{Sim}(\mathbf{f}_k^v, \mathbf{f})}{\tau}\right), \\
\mathcal{L}^{vv}_{neighbor} &= - \frac{1}{N_v} \sum_{k=1}^{N_v} \log \frac{S^v_{k,r}}{S^v_{k,r} + S^v_{k,w}},
\end{align}

\noindent where $\tau$ is a temperature hyperparameter that scales the similarities. By minimizing this contrastive loss, the model learns to maximize the similarities between the instance features and their positive samples while minimizing the similarities with their negative samples.

Similarly, the contrastive loss for infrared-infrared $\mathcal{L}^{rr}_{neighbor}$, infrared-visible $\mathcal{L}^{rv}_{neighbor}$, and visible-infrared $\mathcal{L}^{vr}_{neighbor}$ can be obtained by similar ways. The final optimization for neighbor contrastive learning is denoted by the following combination:
\begin{equation}
\small
\mathcal{L}_{neighbor}=\mathcal{L}^{vv}_{neighbor}+\mathcal{L}^{rr}_{neighbor}+\mathcal{L}^{rv}_{neighbor}+\mathcal{L}^{vr}_{neighbor}.
\end{equation}

\subsection{Multi-Center Contrastive Learning}
Since hierarchical identity information has been obtained, it is intuitive to utilize the centers of multi-granularity clusters as reference points for constructing positive and negative sample sets in contrastive learning. For each visible instance feature, we calculate its cosine similarity with all visible fine-grained cluster features to find the one with the highest similarity. The cosine similarity between the instance feature $\mathbf{f}_k^{v}$ and the fine-grained cluster feature $\mathbf{u}^{vf}_{ij}$ is computed as follows:
\begin{equation}
\text{Sim}(\mathbf{f}_k^{v}, \mathbf{u}^{vf}_{ij}) = \frac{\mathbf{f}_k^{v} \cdot \mathbf{u}^{vf}_{ij}}{\|\mathbf{f}_k^{v}\|\|\mathbf{u}^{vf}_{ij}\|}.
\end{equation}

The fine-grained cluster feature with the highest similarity for a given instance feature $\mathbf{f}_k^v$ can be identified as:
\begin{equation}
\tilde{\mathbf{u}}^v_k = \underset{\mathbf{u}_{ij}^{vf} \in \mathcal{U}^{vf}}{\arg\max}\, \text{Sim}(\mathbf{f}_k^v, \mathbf{u}^{vf}_{ij}).
\end{equation}

Next, we select all fine-grained cluster features within the coarse-grained cluster of $\tilde{\mathbf{u}}^v_k$ as the positive sample set $\mathcal{P}_k^v=\{\mathbf{u}^{vf}_{sj}\}_{j=1}^K$ for the instance feature $\mathbf{f}_k^v$. 
Here, $s$ denotes the coarse-grained pseudo-label of the selected fine-grained feature $\tilde{\mathbf{u}}^v_k$. For the remaining visible coarse-grained clusters, we choose the most similar fine-grained cluster feature within each coarse-grained cluster as the negative sample set $\mathcal{N}_k^v$ for the instance feature $\mathbf{f}_k^v$, which can be represented as:
\begin{equation}
    \mathcal{N}_k^v = \bigcup_{1 \leq i \leq M_v \land i \neq s}\left\{\arg\max_{\mathbf{u}^{vf}_{ij} \in \mathcal{U}^{vf}} \text{Sim}(\mathbf{f}_k^v, \mathbf{u}^{vf}_{ij}) \right\}.
\end{equation}

By obtaining the positive and negative sample sets, the contrastive loss for visible-visible $\mathcal{L}^{vv}_{mccl}$ is defined as follows:
\begin{align}
S^v_{k,p} &= \sum_{\mathbf{u} \in \mathcal{P}_k^v} \exp\left(\frac{\text{Sim}(\mathbf{f}_k^v, \mathbf{u})}{\tau}\right), \\
S^v_{k,n} &= \sum_{\mathbf{u} \in \mathcal{N}_k^v} \exp\left(\frac{\text{Sim}(\mathbf{f}_k^v, \mathbf{u})}{\tau}\right), \\
\mathcal{L}^{vv}_{mccl} &= - \frac{1}{N_v} \sum_{k=1}^{N_v} \log \frac{S^v_{k,p}}{S^v_{k,p} + S^v_{k,n}},
\end{align}

\noindent where $\tau$ is a temperature hyperparameter that scales the similarities. By minimizing this contrastive loss, the model learns to maximize the similarities between the instance features and their positive samples while minimizing the similarities with their negative samples.  

Similarly, the contrastive loss for infrared-infrared $\mathcal{L}^{rr}_{mccl}$, infrared-visible $\mathcal{L}^{rv}_{mccl}$, and visible-infrared $\mathcal{L}^{vr}_{mccl}$ can be obtained by similar ways. The final optimization for multi-center contrastive learning is denoted by the following combination:
\begin{equation}
\mathcal{L}_{mccl} = \mathcal{L}^{vv}_{mccl} + \mathcal{L}^{rr}_{mccl} + \mathcal{L}^{rv}_{mccl} + \mathcal{L}^{vr}_{mccl}.
\end{equation}

The overall loss is denoted as:
\begin{equation}
\mathcal{L}_{total} = \mathcal{L}_{id} + \lambda_1\mathcal{L}_{neighbor}
 + \lambda_2\mathcal{L}_{mccl},
\end{equation}

\noindent where $\lambda_1$ and $\lambda_2$ are hyperparameters to balance the loss terms. Through this optimization, we enable the model to refine representations, thereby improving intra-modal clustering and minimizing cross-modal discrepancies. The training process in one epoch is illustrated in Algorithm~\ref{alg:training}.

\begin{algorithm}[t]  
  \caption{Training process in one epoch}  
  \label{alg:training}  
    \textbf{Require: \\}{Network $f_{\theta}$; 
    Current epoch number $epoch$;\\
    Iterations per epoch $T_t$.\\}
    \textbf{Input: \\}{Visible training dataset ${\mathcal{X}}^v = \left\{\mathbf{x}^v_k \right\}_{k=1}^{N_v}$;\\ Infrared training dataset ${\mathcal{X}}^r = \left\{\mathbf{x}^r_k \right\}_{k=1}^{N_r}$.\\}
    \textbf{Output:\\} Updated network $f_\theta$.\\
    \textbf{1:}
    Extract visible features ${\mathcal{F}}^v = \left\{ \mathbf{f}^v_k \right\}_{k=1}^{N_v}$ and infrared features ${\mathcal{F}}^r = \left\{ \mathbf{f}^r_k \right\}_{k=1}^{N_r}$;\\
    \textbf{2:}
    Cluster features by DBSCAN algorithm to obtain the coarse-grained pseudo-labels ${\mathcal{Y}}^{vc} = \left\{\mathbf{y}^{vc}_i\right\}_{i=1}^{M_v}$ and ${\mathcal{Y}}^{rc} = \left\{\mathbf{y}^{rc}_i\right\}_{i=1}^{M_r}$;\\
    \textbf{3:}
    Initialize the coarse-grained cluster features ${\mathcal{U}}^{vc} = \left\{\mathbf{u}^{vc}_i\right\}_{i=1}^{M_v}$ and ${\mathcal{U}}^{rc} = \left\{\mathbf{u}^{rc}_i\right\}_{i=1}^{M_r}$ by cluster centroids;\\
    \textbf{4:}
    Cluster features within the same coarse-grained pseudo-label by K-means algorithm to obtain the fine-grained pseudo-labels ${\mathcal{Y}}^{vf}=\left\{\mathbf{y}^{vf}_{ij}\right\}_{i=1,j=1}^{i=M_v,j=K}$ and ${\mathcal{Y}}^{rf}=\left\{\mathbf{y}^{rf}_{ij}\right\}_{i=1,j=1}^{i=M_r,j=K}$;\\
    \textbf{5:}
    Initialize the fine-grained cluster features $\mathcal{U}^{vf} = \left\{\mathbf{u}^{vf}_{ij} \right\}_{i=1,j=1}^{i=M_v,j=K}$  and $\mathcal{U}^{rf} = \left\{\mathbf{u}^{rf}_{ij} \right\}_{i=1,j=1}^{i=M_r,j=K}$  by cluster centroids;\\
    \textbf{6:}
    \If{$epoch \% 2 == 0$}{
        Obtain the modality-unified pseudo-labels ${\mathcal{\hat{Y}}}^{v} = \left\{\mathbf{\hat{y}}^{v}_i\right\}_{i=1}^{M_v}$ by BRST mechanism;\\}
    \If{$epoch \% 2 == 1$}{
        Obtain the modality-unified pseudo-labels ${\mathcal{\hat{Y}}}^{r} = \left\{\mathbf{\hat{y}}^{r}_i\right\}_{i=1}^{M_r}$ by BRST mechanism;\\
    }
    \textbf{7:}
    \For{$t$ \textbf{in} [1, $T_t$]}{
        Compute $\mathcal{L}_{id}$ as Eq.\textcolor{blue}{7};\\
        Compute $\mathcal{L}_{neighbor}$ as Eq.\textcolor{blue}{17};\\
        Compute $\mathcal{L}_{mccl}$ as Eq.\textcolor{blue}{24};\\
        Compute the total loss $\mathcal{L}_{total}$ as Eq.\textcolor{blue}{25};\\
        Update the coarse-grained cluster features;\\
        Update the fine-grained cluster features;\\
        Update the network parameters $\theta$.\\
    }
    
\end{algorithm}

\subsection{Bidirectional Reverse Selection Transmission}
We introduce the Bidirectional Reverse Selection Transmission (BRST) mechanism to ensure robust and accurate label assignment for visible instance features by utilizing the coarse-grained cluster features of the infrared modality, and vice versa. These features serve as comprehensive representations in the hierarchical structure by encapsulating fine-grained variations, making them effective for achieving reliable identity alignment across modalities.

Specifically, for each visible instance feature, we calculate its cosine similarity with all infrared coarse-grained cluster features. This computation yields a similarity matrix $\mathbf{S} \in \mathbb{R}^{N_v \times M_r}$, where each row corresponds to one of the $N_v$ visible instances, and each column corresponds to one of the $M_r$ infrared coarse-grained clusters. Each element $\mathbf{S}_{ki}$ in the matrix represents the similarity score between the $k$-th visible instance feature and the $i$-th infrared cluster feature. 

To assign labels to visible instances, we first calculate the maximum similarity value and its corresponding column for each visible instance feature $\mathbf{f}_k^v$, which is defined as:
\begin{equation}
    m_r = \max_{i} \mathbf{S}_{ki}, \,
    i^{*} = \arg\max_{i} \mathbf{S}_{ki},
\end{equation}

\noindent where $m_r$ represents the highest similarity value in row $k$ of the similarity matrix, and $i^*$ denotes the column of the corresponding infrared cluster.
To ensure bidirectional consistency, a reverse selection step is performed. Specifically, for the cluster linked by $i^*$, we identify the maximum similarity value in the corresponding column:
\begin{equation}
    m_c = \max_{k} \mathbf{S}_{ki^{*}},
\end{equation}

\noindent where $m_c$ denotes the maximum similarity value in column $i^*$ of the similarity matrix. 

The final label assignment for the $k$-th visible instance is determined based on a comparison between the two maximum similarity values $m_r$ and $m_c$. We define that if $m_r > \gamma \cdot m_c$, the visible instance is assigned the label corresponding to column $i^*$. Here, the parameter $\gamma$ serves as a critical threshold that controls the selection criteria, ensuring that only sufficiently confident matches are retained. The set of modality-unified pseudo-labels is defined as ${\mathcal{\hat{Y}}}^{v} = \left\{\mathbf{\hat{y}}^{v}_i\right\}_{i=1}^{M_v}$.

For each infrared instance feature, we perform the same operations for filtering and matching, thus ensuring a bidirectional approach. Similarly, we can obtain the set of modality-unified pseudo-labels ${\mathcal{\hat{Y}}}^{r} = \left\{\mathbf{\hat{y}}^{r}_i\right\}_{i=1}^{M_r}$. This mechanism effectively addresses model stagnation by alternately transferring labels between visible and infrared instances through multi-perspective transformations. To ensure the reliability of these transfers, this mechanism filters out unreliable pseudo-label matches, thereby enhancing the quality and robustness.

\section{Experiment}

\begin{table*}[!htbp]
    \centering
    \caption{Comparison with the state-of-the-art methods on SYSU-MM01 and RegDB. ``GUR$^\ast$" denotes GUR without camera labels. Since our method does not require any camera label information, we do not report the results of GUR with camera labels for fair comparison.}
    \footnotesize
    \resizebox{\textwidth}{!}{
        \begin{tabular}{l|l|ccc|ccc|ccc|ccc}
            \hline
            \multirow{3}{*}{Methods} & \multirow{3}{*}{Venue} & \multicolumn{6}{c|}{SYSU-MM01} & \multicolumn{6}{c}{RegDB}\\
            \cline{3-14}
            & & \multicolumn{3}{c|}{All Search} & \multicolumn{3}{c|}{Indoor Search} & \multicolumn{3}{c|}{Visible-to-Infrared} & \multicolumn{3}{c}{Infrared-to-Visible}\\ 
            \cline{3-14}
            & & R1 & mAP & mINP & R1 & mAP & mINP & R1 & mAP & mINP & R1 & mAP & mINP\\
            \hline
            \multicolumn{12}{l}{\textit{Supervised VI-ReID methods}} \\ 
            \hline
            SPOT ~\cite{chen2022structure} & TIP-22 & 65.34 & 62.25 & 48.86 & 69.42 & 74.63 & 70.48 & 80.35 & 72.46 & 56.19 & 79.37 & 72.26 & 56.06 \\
            DART~\cite{Yang_2022_CVPR} & CVPR-22 & 68.72 & 66.29 & - & 72.52 & 78.17 & - & 83.60 & 75.67 & - & 81.97 & 73.78 & - \\
            FMCNet~\cite{zhang2022fmcnet} & CVPR-22 & 66.34 & 62.51 & - & 68.15 & 74.09 & - & 89.12 & 84.43 & - & 88.38 & 83.86 & - \\
            MAUM~\cite{liu2022learning} & CVPR-22 & 71.68 & 68.79 & - & 76.97 & 81.94 & - & 87.87 & 85.09 & - & 86.95 & 84.34 & - \\ 
            TransVI ~\cite{chai2023dual} & TCSVT-23 & 71.36 & 68.63 & - & 77.40 & 81.31 & - & 96.66 & 91.22 & - & 96.30 & 91.21 & - \\
            DEEN~\cite{Zhang_2023_CVPR} & CVPR-23 & 74.70 & 71.80 & - & 80.30 & 83.30 & - & 91.10 & 85.10 & - & 89.50 & 83.40 & - \\
            CAL~\cite{Wu_2023_ICCV} & ICCV-23 & 74.66 & 71.73 & - & 79.69 & 83.68 & - & 94.51 & 88.67 & - & 93.64 & 87.61 & - \\
            SAAI~\cite{Fang_2023_ICCV} & ICCV-23 & 75.90 & 77.03 & - & 83.20 & 88.01 & - & 91.07 & 91.45 & - & 92.09 & 92.01 & - \\
            PMCM~\cite{qian2025visible} & IJCAI-23 & 75.54 & 71.16 & - & 81.52 & 84.33 & - & 93.09 & 89.57 & - & 91.44 & 87.15 & - \\ 
            STAR ~\cite{wu2023style} & TMM-23 & 76.07 & 72.73 & - & 83.47 & 85.76 & - & 94.09 & 88.75 & - & 93.30 & 88.20 & - \\
            PartMix~\cite{kim2023partmix} & CVPR-23 & 77.78 & 74.62 & - & 81.52 & 84.38 & - & 84.93 & 82.52 & - & 85.66 & 82.27 & - \\ 
            SEFL~\cite{Feng_2023_CVPR} & CVPR-23 & 77.12 & 72.33 & - & 82.07 & 82.95 & - & 95.35 & 89.98 & - & 97.57 & 91.41 & - \\ 
            IDKL~\cite{ren2024implicit} & CVPR-24 & 81.42 & 79.85 & - & 87.14 & 89.37 & - & 94.72 & 90.19 & - & 94.22 & 90.43 & - \\ 
            HOS-Net~\cite{qiu2024high} & AAAI-24 & 75.60 & 74.20 & - & 84.20 & 86.70 & - & 94.70 & 90.40 & - & 93.30 & 89.20 & - \\
            DMA~\cite{cui2024dma} & TIFS-24 & 74.57 & 70.41 & 56.50 & 82.85 & 85.10 & - & 93.30 & 88.34 & - & 91.50 & 86.80 & - \\ 
            \hline 
            \multicolumn{12}{l}{\textit{Semi-supervised VI-ReID methods}} \\ 
            \hline
            OTLA~\cite{OTLA} & ECCV-22 & 48.20 & 43.90 & - & 47.40 & 56.80 & - & 49.90 & 41.80 & - & 49.60 & 42.80 & - \\ 
            TAA~\cite{yang2023translation} & TIP-23 & 48.77 & 42.43 & 25.37 & 50.12 & 56.02 & 49.96 & 62.23 & 56.00 & 41.51 & 63.79 & 56.53 & 38.99 \\ 
            DPIS~\cite{Shi_2023_ICCV} & ICCV-23 & 58.40 & 55.60 & - & 63.00 & 70.00 & - & 62.30 & 53.20 & - & 61.50 & 52.70 & -\\
            \hline
            \multicolumn{12}{l}{\textit{Unsupervised VI-ReID methods}} \\ 
            \hline
            ADCA~\cite{adca} & MM-22 & 45.51 & 42.73 & 28.29 & 50.60 & 59.11 & 55.17 & 67.20 & 64.05 & 52.67 & 68.48 & 63.81 & 49.62 \\
            CHCR~\cite{CHCR} & TCSVT-23 & 47.72 & 45.34 & - & 50.12 & 42.17 & - & 68.18 & 63.75 & - & 69.08 & 63.95 & - \\
            MBCCM~\cite{MBCCM} & MM-23 & 53.14 & 48.16 & 32.41 & 55.21 & 61.98 & 57.13 & 83.79 & 77.87 & 65.04 & 82.82 & 76.74 & 61.73 \\
            DOTLA~\cite{DOTLA} & MM-23 & 50.36 & 47.36 & 32.40 & 53.47 & 61.73 & 57.35 & 85.63 & 76.71 & 61.58 & 82.91 & 74.97 & 58.60 \\
            CCLNet~\cite{CCLNet} & MM-23 & 54.03 & 50.19 & - & 56.68 & 65.12 & - & 69.94 & 65.53 & - & 70.17 & 66.66 & - \\
            PGM~\cite{PGM} & CVPR-23 & 57.27 & 51.78 & 34.96 & 56.23 & 62.74 & 58.13 & 69.48 & 65.41 & - & 69.85 & 65.17 & - \\
            GUR$^\ast$~\cite{Yang_2023_ICCV} & ICCV-23 & 60.95 & 56.99 & 41.85 & 64.22 & 69.49 & 64.81 & 73.91 & 70.23 & 58.88 & 75.00 & 69.94 & 56.21 \\
            MIMR~\cite{pang2024mimr} & KBS-24 & 46.56 & 45.88 & - & 52.26 & 60.93 & - & 68.76 & 64.33 & - & 68.76 & 63.83 & - \\
            SCA-RCP~\cite{li2024inter} & TKDE-24 & 51.41 & 48.52 & 33.56 & 56.77 & 64.19 & 59.25 & 85.59 & 78.12 & - & 82.41 & 75.73 & - \\
            BCGM~\cite{teng2024enhancing} & MM-24 & 61.70 & 56.10 & 38.70 & 60.90 & 66.50 & 62.30 & 86.80 & 81.70 & 68.60 & 86.70 & 82.30 & 71.10 \\
            RPNR~\cite{yin2024robust} & MM-24 & 65.20 & 60.00 & - & 68.90 & 74.40 & - & 90.90 & 84.70 & - & 90.10 & 83.20 & - \\
            MMM~\cite{shi2025multi} & ECCV-24 & 61.60 & 57.90 & - & 64.40 & 70.40 & - & 89.70 & 80.50 & - & 85.80 & 77.00 & - \\
            PCLHD~\cite{shi2024learning} & NIPS-24 & 64.40 & 58.70 & - & 69.50 & 74.40 & - & 84.30 & 80.70 & - & 82.70 & 78.40 & - \\
            MULT~\cite{he2024exploring} & IJCV-24 & 65.03 & 58.62 & 42.77 & 65.35 & 71.24 & 66.60 & 91.50 & 83.73 & 69.13 & 89.08 & 80.88 & 64.03 \\
            SDCL~\cite{yang2024shallow} & CVPR-24 & 64.49 & 63.24 & 51.06 & 71.37 & 76.90 & 73.50 & 86.91 & 78.92 & 62.83 & 85.76 & 77.25 & 59.57 \\
            IMSL~\cite{pang2024inter} & TCSVT-24 & 57.96 & 53.93 & - & 58.30 & 64.31 & - & 70.08 & 66.30 & - & 70.67 & 66.35 & - \\
            PCAL~\cite{yang2025progressive} & TIFS-25 & 57.94 & 52.85 & 36.90 & 60.07 & 66.73 & 62.09 & 86.43 & 82.51 & 72.33 & 86.21 & 81.23 & 68.71 \\
            SALCR~\cite{cheng2025semantic} & IJCV-25 & 64.44 & 60.44 & 45.19 & 67.17 & 72.88 & 68.73 & 90.58 & 83.87 & 70.76 & 88.69 & 82.66 & 66.89 \\
            \hline
            Ours & - & \textbf{66.30} & \textbf{64.95} & \textbf{52.62} & \textbf{71.81} & \textbf{77.52} & \textbf{74.25} & \textbf{92.82} & \textbf{86.61} & \textbf{73.69} & \textbf{92.24} & \textbf{85.43} & \textbf{70.87} \\
            \hline    
        \end{tabular}
    }
\label{tab:sota_comparison}
\end{table*}

\subsection{Datasets and Evaluation Protocol} 

\paragraph{Datasets} We evaluate the proposed HIL framework on two widely-used visible-infrared person ReID datasets, namely SYSU-MM01 ~\cite{SYSU-MM01} and RegDB ~\cite{RegDB}. The SYSU-MM01 dataset is collected by 6 different cameras (4 RGB cameras and 2 IR cameras), including 287,628 RGB images and 15,792 IR images of 491 identities. The RegDB dataset is captured by a dual-camera system that aligns visible and infrared images. It includes 412 identities, and each has 10 infrared images and 10 visible images. 

\paragraph{Evaluation Protocol} Cumulative Matching Characteristics (CMC), mean Average Precision (mAP), and mean Inverse Negative Penalty (mINP) ~\cite{ye2021deep} are adopted as the evaluation metrics. For the RegDB dataset, we randomly select 206 identities for training and use the remaining 206 identities for testing, evaluating the method in Visible-to-Infrared mode and Infrared-to-Visible mode. We calculate the average result obtained from 10 random splits of the training set and testing set. For the SYSU-MM01 dataset, the training set contains 22,258 visible images and 11,909 infrared images of 395 identities. The remaining 96 identities are adopted for testing, containing 3,803 infrared images for the query set and 301 randomly selected visible images for the gallery set, evaluating the method in All Search and Indoor Search modes. We also calculate the average result obtained from 10 random gallery set selections.

\subsection{Implementation Details}

The proposed HIL framework is implemented using PyTorch, with the feature extractor from TransReID~\cite{He_2021_ICCV} serving as the backbone network and augmented with dual contrastive learning~\cite{adca}. Coarse-grained pseudo-labels are generated at the start of each training epoch using DBSCAN~\cite{ester1996density}, followed by the secondary clustering for the generation of fine-grained pseudo-labels using K-means~\cite{macqueen1967some}. The learning rate is initialized to 0.000035 and reduced by 0.1 every 20 epochs. The images are resized to 288 × 144 before being fed into the network. Each training batch consists of 8 pseudo-identities, with 16 instances sampled per pseudo-identity for each modality. $K$ and $\gamma$ are set to 9 and 0.5 for SYSU-MM01 and 2 and 0.8 for RegDB. The temperature factor $\tau$ is set to 0.05 and the momentum updating factor $\mu$ is set to 0.1. Before standard training, we train the SDCL~\cite{yang2024shallow} framework for the initial 30 epochs. Subsequently, we continue training our framework for 30 epochs based on this pre-trained model, using SGD as the optimizer.

\begin{table*}[!htbp]
    \centering
    \caption{Ablation study on the individual components of our method on SYSU-MM01 and RegDB.}
    \footnotesize
    \resizebox{\textwidth}{!}{
        \begin{tabular}{c|ccc|ccc|ccc|ccc|ccc}
            \hline
            \multirow{3}{*}{ID} &
            \multicolumn{3}{c|}{Components} &
            \multicolumn{6}{c|}{SYSU-MM01} &
            \multicolumn{6}{c}{RegDB}\\
            \cline{2-4} \cline{5-16}
            & \multirow{2}{*}{Baseline} & \multirow{2}{*}{MCCL} & \multirow{2}{*}{BRST}
            & \multicolumn{3}{c|}{All Search}
            & \multicolumn{3}{c|}{Indoor Search} 
            & \multicolumn{3}{c|}{Visible-to-Infrared} 
            & \multicolumn{3}{c}{Infrared-to-Visible}\\ 
            \cline{5-16}
            & & & & R1 & mAP & mINP & R1 & mAP & mINP & R1 & mAP & mINP & R1 & mAP & mINP\\
            \hline
            1 & \checkmark &  &  & 54.88 & 52.62 & 38.06 & 59.02 & 66.14 & 61.85 & 84.11 & 70.44 & 48.45 & 81.66 & 67.94 & 45.31 \\
            2 & \checkmark & \checkmark &  & 55.83 & 53.76 & 39.36 & 60.67 & 67.62 & 63.39 & 87.16 & 75.94 & 56.36 & 85.05 & 73.50 & 52.92 \\
            3 & \checkmark &  & \checkmark & 65.53 & 63.60 & 50.67 & 71.06 & 76.54 & 73.08 & 92.23 & 85.34 & 70.88 & 91.52 & 84.12 & 68.07 \\
            4 & \checkmark & \checkmark & \checkmark & \textbf{66.30} & \textbf{64.95} & \textbf{52.62} & \textbf{71.81} & \textbf{77.52} & \textbf{74.25} & \textbf{92.82} & \textbf{86.61} & \textbf{73.69} & \textbf{92.24} & \textbf{85.43} & \textbf{70.87}  \\
            \hline
        \end{tabular}
    }
    \label{tab:ablation_line}
\end{table*}

\subsection{Comparison with State-of-the-art Methods}

To comprehensively evaluate our method in the VI-ReID task, we compare our method with 15 supervised methods, 3 semi-supervised methods, and 18 unsupervised methods. The comparison results on the SYSU-MM01 and RegDB datasets are reported in Table \ref{tab:sota_comparison}.

\paragraph{Comparison with Supervised VI-ReID Methods} Our proposed HIL method demonstrates competitive performance compared to the supervised method FMCNet~\cite{zhang2022fmcnet} on the SYSU-MM01 dataset and achieves performance close to state-of-the-art supervised methods on RegDB. Although supervised methods benefit from precise manual annotations, our approach achieves comparable results, showcasing its robustness without relying on labeled data.

\paragraph{Comparison with Semi-Supervised VI-ReID Methods} Semi-supervised VI-ReID methods aim to minimize labeling costs by utilizing partially labeled data while maintaining performance. In the realm of existing semi-supervised methods, our proposed approach surpasses the three main methods reported in the literature, achieving state-of-the-art performance without relying on any manual annotations. By eliminating the need for labeled data, our method demonstrates its ability to significantly reduce the dependency on human annotation efforts, while still achieving competitive and robust results across challenging cross-modal VI-ReID tasks.

\paragraph{Comparison with Unsupervised VI-ReID Methods} Our method outperforms state-of-the-art unsupervised VI-ReID approaches by a considerable margin. Specifically, it achieves 64.95\% mAP and 66.30\% Rank-1 on SYSU-MM01 (All Search) and 86.61\% mAP and 92.82\% Rank-1 on RegDB (Visible-to-Infrared). Compared to the best-performing unsupervised method, SDCL~\cite{yang2024shallow}, our method exceeds it by 1.81\% on SYSU-MM01 (All Search) and 5.91\% on RegDB (Visible-to-Infrared) in Rank-1, and by 1.71\% on SYSU-MM01 (All Search) and 7.69\% on RegDB (Visible-to-Infrared) in mAP. These results underscore the effectiveness of our approach in building robust cross-modal relationships and addressing noise in cross-modal label correspondences.

\begin{table*}[!htbp]
    \centering
    \caption{Comparison of cross-modal label association methods for unsupervised VI-ReID on SYSU-MM01 and RegDB.}
    \footnotesize
    \resizebox{\textwidth}{!}{
        \begin{tabular}{c|c|ccc|ccc|ccc|ccc}
            \hline
            \multirow{3}{*}{ID} & 
            \multirow{3}{*}{Methods} & 
            \multicolumn{6}{c|}{SYSU-MM01} &
            \multicolumn{6}{c}{RegDB} \\
            \cline{3-14}
            & & \multicolumn{3}{c|}{All Search} & 
            \multicolumn{3}{c|}{Indoor Search} &
            \multicolumn{3}{c|}{Visible-to-Infrared} & \multicolumn{3}{c}{Infrared-to-Visible} \\
            \cline{3-14}
            & & R1 & mAP & mINP & R1 & mAP & mINP & R1 & mAP & mINP & R1 & mAP & mINP\\
            \hline
            1 & \multicolumn{1}{l|}{Baseline} & 54.88 & 52.62 & 38.06 & 59.02 & 66.14 & 61.85 & 84.11 & 70.44 & 48.45 & 81.66 & 67.94 & 45.31\\
            2 & \multicolumn{1}{l|}{Baseline + OPTM~\cite{yin2024robust}} & 64.31 & 62.88 & 50.34 & 70.27 & 76.00 & 72.51 & 91.33 & 84.31 & 70.04 & 90.76 & 83.32 & 67.73\\
            3 & \multicolumn{1}{l|}{Baseline + CRA~\cite{yang2024shallow}} & 65.04 & 63.35 & 50.51 & 70.73 & 76.36  & 72.93 & 91.82 & 84.80 & 70.63 & 91.04 & 83.75 & 67.89 \\
            4 & \multicolumn{1}{l|}{Baseline + BRST (Ours)} & \textbf{65.53} & \textbf{63.60} & \textbf{50.67} & \textbf{71.06} & \textbf{76.54} & \textbf{73.08} & \textbf{92.23} & \textbf{85.34} & \textbf{70.88} & \textbf{91.52} & \textbf{84.12} & \textbf{68.07} \\
            \hline
        \end{tabular}
        \label{3}
    }
\end{table*}

\begin{figure}[!t]
    \centering
    \includegraphics[width=\linewidth]{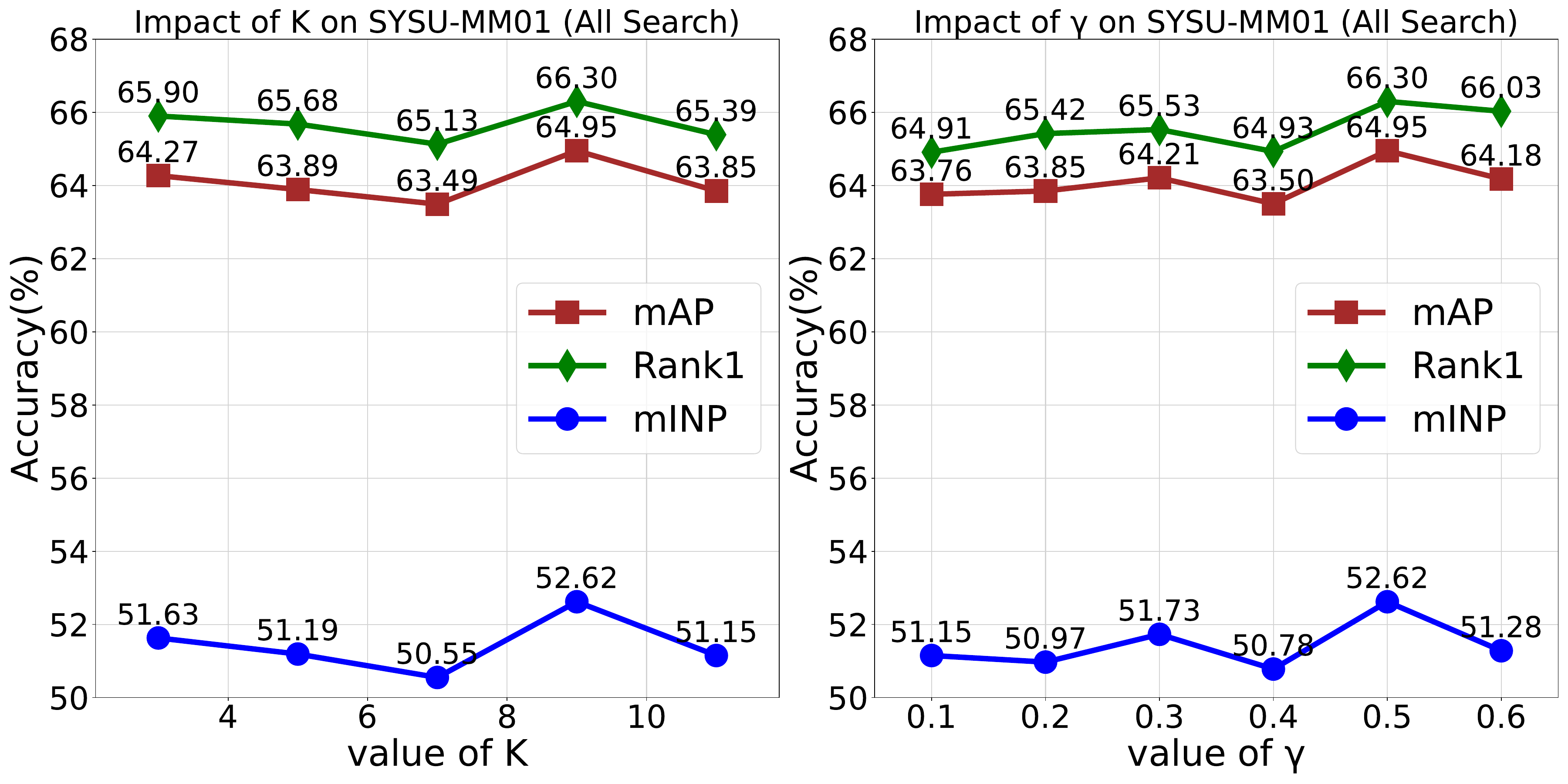}
    \caption{Hyperparameter analysis of $K$ and $\gamma$ on SYSU-MM01.}
    \label{fig:3}
\end{figure}

\begin{figure}[!t]
    \centering
    \includegraphics[width=\linewidth]{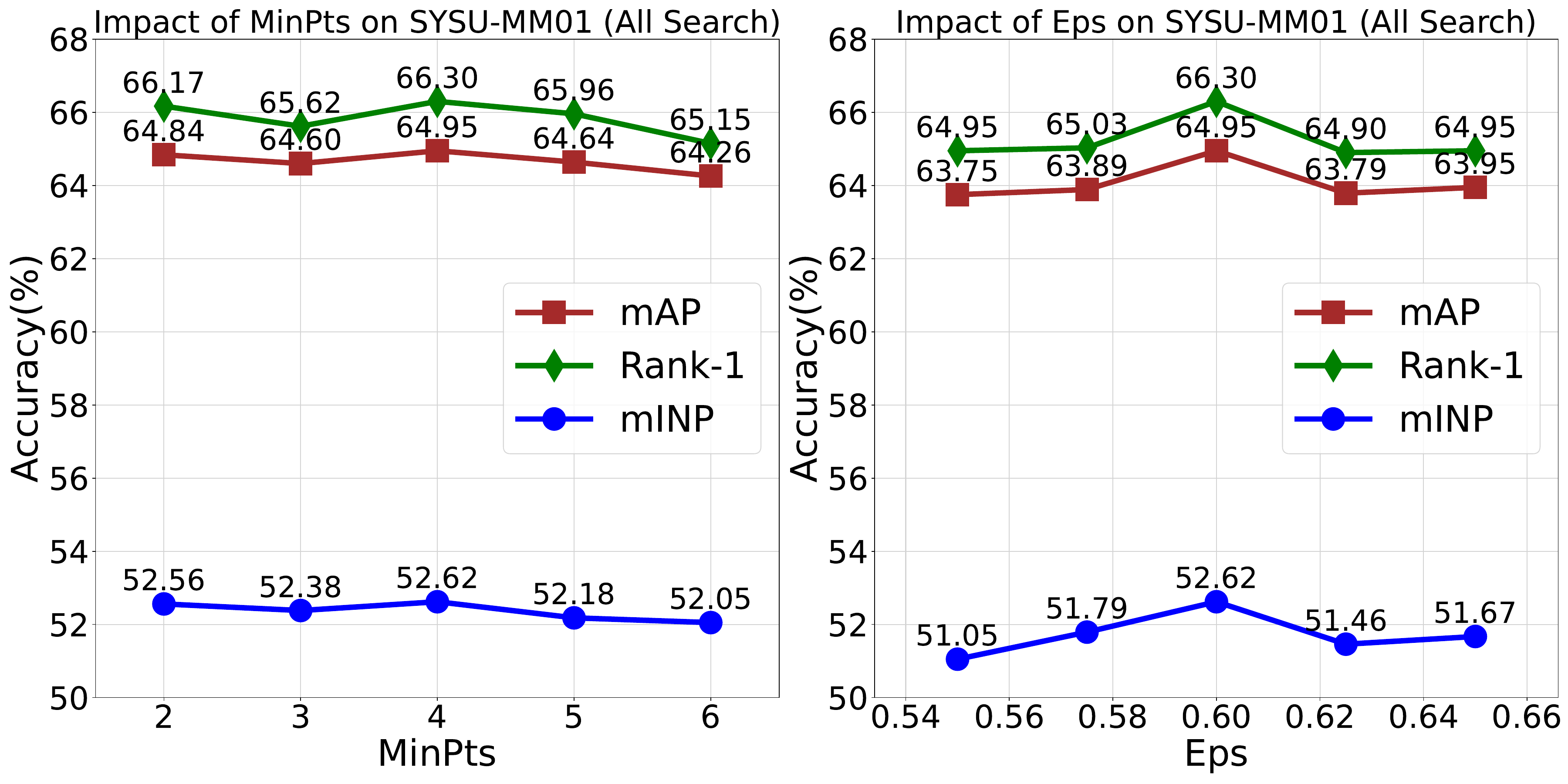}
    \caption{Hyperparameter analysis of MinPts and Eps on SYSU-MM01.}
    \label{fig:6}
\end{figure}

\begin{figure}[!t]
    \centering
    \includegraphics[width=\linewidth]{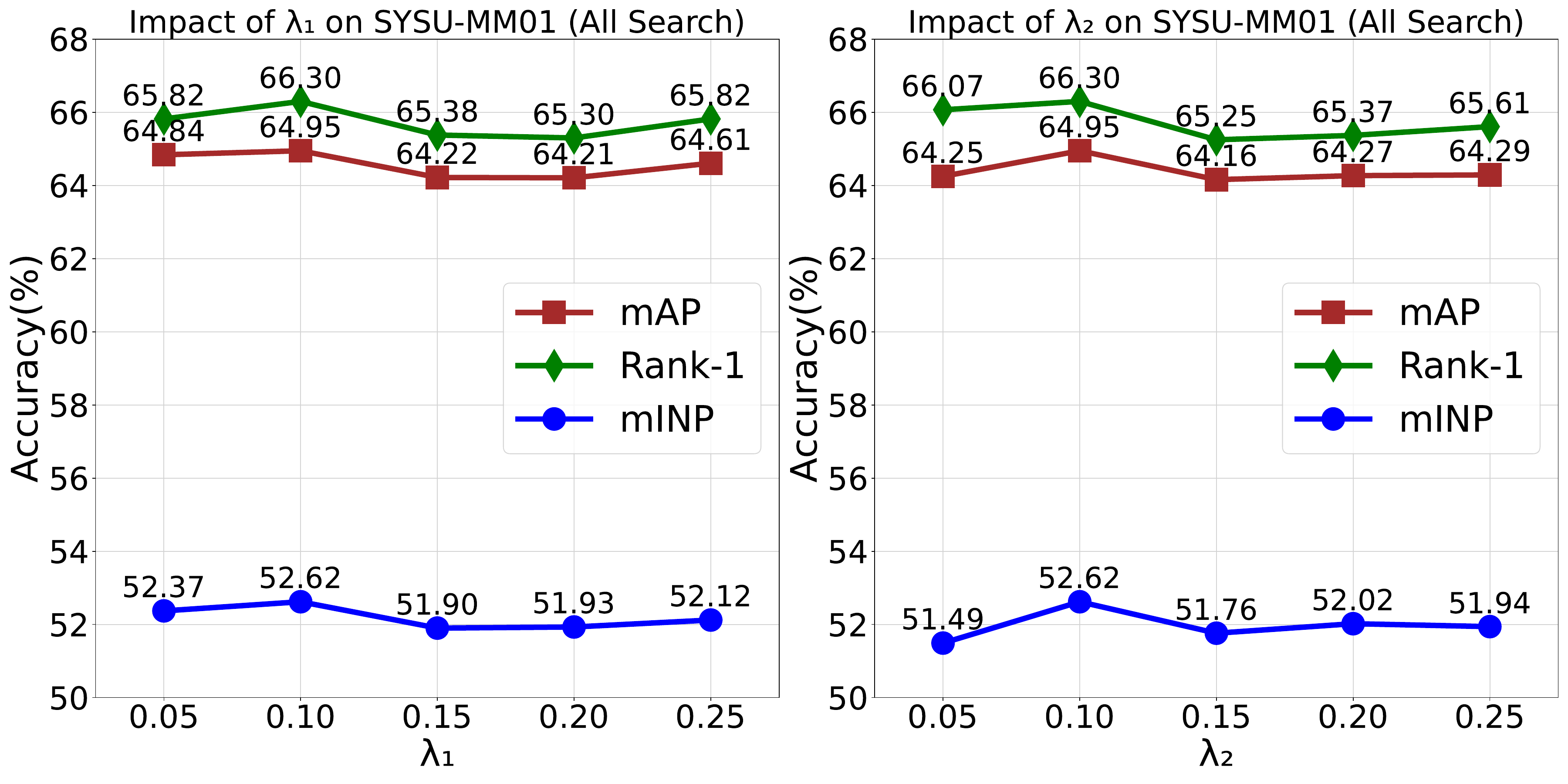}
    \caption{Hyperparameter analysis of $\lambda_1$ and $\lambda_2$ on SYSU-MM01.}
    \label{fig:9}
\end{figure}

\begin{figure*}[!t]
    \centering
    \includegraphics[width=\linewidth]{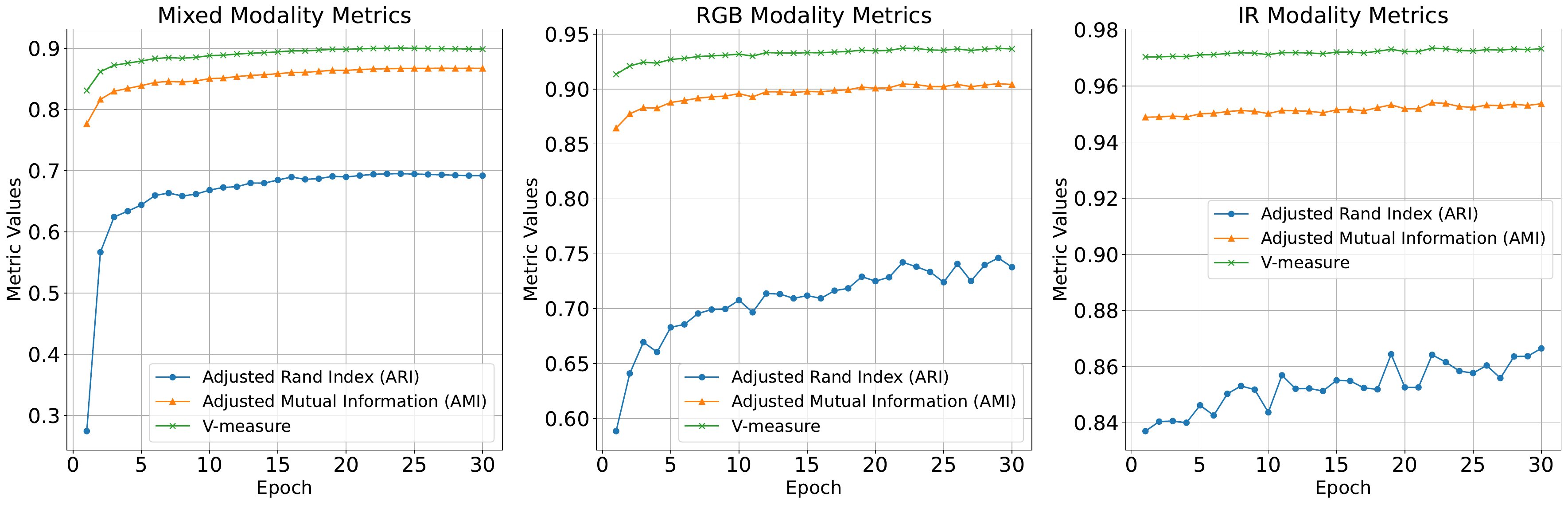}
    \caption{The variation of clustering quality produced by DBSCAN on the SYSU-MM01 dataset with epoch. The RGB modality represents the clustering quality of pseudo-labels generated using only visible light image modality data. The IR modality indicates the clustering quality of pseudo-labels generated using only infrared image modality data. The mixed modality refers to the overall pseudo-label clustering quality formed by fusing visible light and infrared image modality data.}
    \label{fig:7}
\end{figure*}

\begin{figure}[!t]
    \centering
    \includegraphics[width=0.98\linewidth]{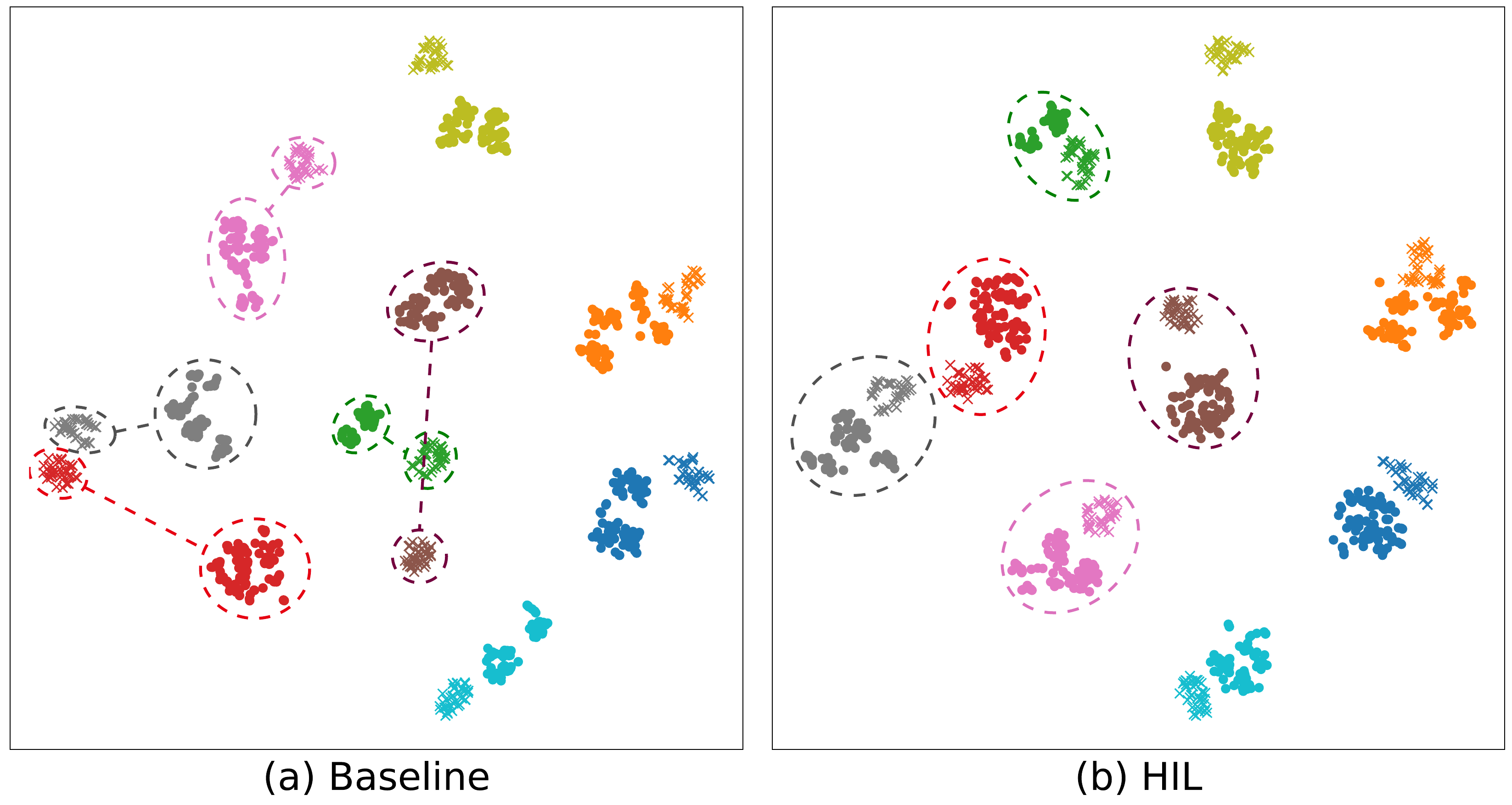}
    \caption{T-SNE visualization of features for randomly selected identities. Specifically, colors indicate identities, with circular markers denoting the visible modality and cross markers indicating the infrared modality.}
    \label{fig:4}
\end{figure}

\begin{figure}[!t]
    \centering
    \includegraphics[width=\linewidth]{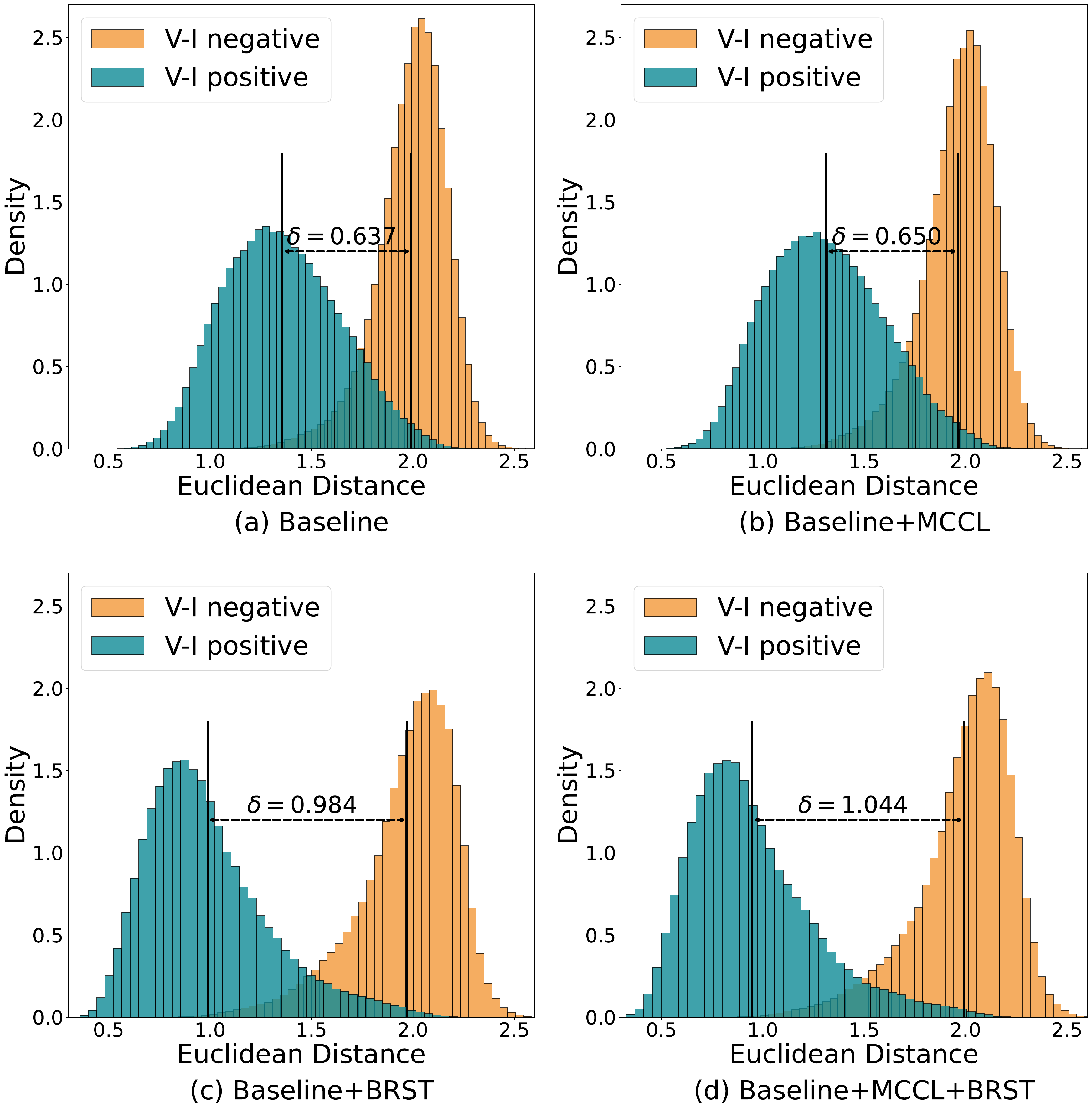}
    \caption{Visualization of Euclidean Distance distribution of randomly selected visual-infrared positive and negative pairs with various combinations. $\delta$ represents the average margin between the distances of positive pairs and negative pairs. As $\delta$ increases, the separability of positive and negative pairs in the cross-modal setting becomes stronger.}
    \label{fig:5}
\end{figure}

\begin{figure*}[!t]
    \centering
    \includegraphics[width=\linewidth]{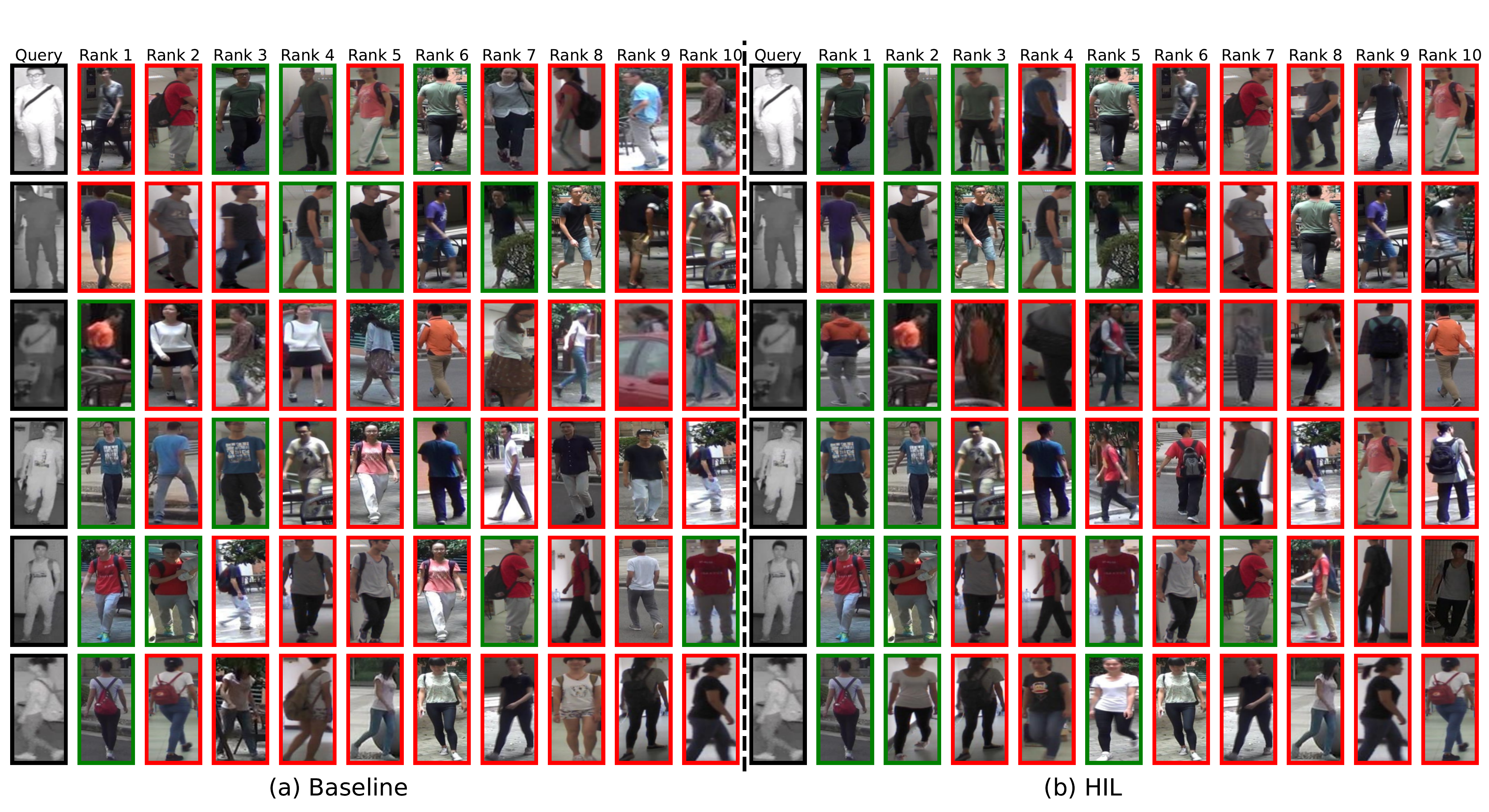}
    \caption{Visualization of the ranking lists on the SYSU-MM01 dataset. The persons who are different from the query persons are marked with red boxes, while those who are the same as the query are marked with green boxes.}
    \label{fig:8}
\end{figure*}

\subsection{Ablation Study}

In this subsection, we conduct ablation experiments to validate the effectiveness of each component in our method. The results are reported in Table \ref{tab:ablation_line}. 

\paragraph{Baseline} We denotes the augmented dual-contrastive learning framework~\cite{adca} with a dual-path transformer architecture. The network optimization is conducted using only the identity loss $\mathcal{L}_{id}$ and the neighbor loss $\mathcal{L}_{neighbor}$ during training.

\paragraph{Effectiveness of MCCL} 

The effectiveness of MCCL is validated through comprehensive comparisons. Specifically, when comparing row 1 and row 2, we observe significant performance improvements on both SYSU-MM01 and RegDB benchmarks in different settings. On SYSU-MM01, MCCL achieves a notable gain of +1.14\%/+0.95\% in mAP/Rank-1, while on RegDB, it leads to an even more substantial improvement of +5.50\%/+3.05\% in mAP/Rank-1 compared to the baseline. Similarly, when comparing row 3 and row 4, the inclusion of MCCL yields further enhancements. For SYSU-MM01 (All Search), MCCL provides an additional boost of +1.35\%/+0.77\% in mAP/Rank-1. On RegDB (Visible-to-Infrared), the improvements are +1.27\%/+0.59\% in mAP/Rank-1. These results clearly demonstrate that the proposed MCCL effectively leverages hierarchical identity information, thereby reducing cross-modal discrepancies and improving the quality of feature representations across modalities.

\paragraph{Effectiveness of BRST} The effectiveness of BRST is demonstrated when comparing row 1 and row 3, with improvements of +10.98\%/+10.65\% mAP/Rank-1 on SYSU-MM01 (All Search) and +14.90\%/+8.12\% mAP/Rank-1 on RegDB (Visible-to-Infrared) compared to the baseline. Additionally, when comparing row 2 and row 4, we observe improvements of +11.19\%/+10.47\% mAP/Rank-1 on SYSU-MM01 (All Search) and +10.67\%/+5.66\% mAP/Rank-1 on RegDB (Visible-to-Infrared) by using BRST. To further evaluate the effectiveness of BRST, we compare BRST with other cross-modal label association methods in Table \ref{3}. The results demonstrate that our BRST provides higher-quality associations than other methods, ensuring robust cross-modal correspondences by filtering unreliable pseudo-label matches through a reverse selection process.

\subsection{Further Analysis} 

\paragraph{Hyperparameter Analysis} 

There are two critical hyperparameters in our method, $K$ and $\gamma$, whose impacts on performance are quantitatively evaluated across a range of values, as shown in Figure \ref{fig:3}. The left panel of the figure illustrates the effect of $K$, where mAP, Rank-1 accuracy, and mINP all reach their highest values when $K$ is set to 9. This setting achieves an optimal balance between cross-modal alignment and noise suppression. Similarly, the right panel highlights the influence of $\gamma$, showing that the best performance is achieved when $\gamma$ is set to 0.5. At this value, the method maintains robust bidirectional selection while avoiding excessively strict or overly lenient matching criteria. These optimal hyperparameter values maximize accuracy across all evaluation metrics, underscoring their significance in enhancing the overall performance of our approach.

In DBSCAN, the two key hyperparameters MinPts and Eps and their impacts on performance have been quantitatively assessed across a range of values, as illustrated in Figure \ref{fig:6}. MinPts is the minimum number of points to form a dense cluster, and Eps is the maximum neighborhood distance. The left panel demonstrates that when MinPts is set to 4, mAP, Rank-1 accuracy, and mINP all reach their highest values, where a smaller MinPts enhances sensitivity to local feature clusters but may introduce noise if too small, while larger values risk losing fine-grained features and reducing accuracy. Similarly, the right panel highlights that the best performance occurs when Eps is set to 0.6, as an appropriate Eps balances the model’s ability to capture valid features and filter interference. These results highlight the critical role of hyperparameter tuning in optimizing clustering performance for DBSCAN.

In terms of loss, our method includes two crucial hyperparameters, $\lambda_1$ and $\lambda_2$, whose impacts on performance are quantitatively evaluated across a range of values, as shown in Figure \ref{fig:9}. The left panel of the figure illustrates the effect of $\lambda_1$, where mAP, Rank-1 accuracy, and mINP all reach their peak values when $\lambda_1$ is set to 0.10. Similarly, the right panel demonstrates that the highest performance across all three metrics is achieved when $\lambda_2$ is set to 0.1. When both $\lambda_1$ and $\lambda_2$ are set to 0.1, these hyperparameters effectively balance the contributions of $\mathcal{L}{id}$, $\mathcal{L}{neighbor}$, and $\mathcal{L}_{mccl}$, thereby enhancing both the discriminative power and generalization ability of the model.

\paragraph{Clustering Quality} 

The overall performance is closely linked to the quality of clustering. As depicted in Fig.\ref{fig:7}, the initial clusters produced by DBSCAN on the SYSU-MM01 dataset exhibit suboptimal performance. However, clustering quality improves significantly throughout iterative training. Concretely, metrics including Adjusted Rand Index (ARI)~\cite{hubert1985comparing}, Adjusted Mutual Information (AMI)~\cite{vinh2009information}, and V-measure~\cite{hirschberg2007v} for all modalities all exhibit an upward trend as the number of training epochs increases. Specifically, MCCL enhances intra-modal clustering by refining feature representations using fine-grained prototypes. At the same time, BRST improves cross-modal clustering by filtering out unreliable pseudo-label matches during label association, leading to more accurate and stable cross-modal alignment.

\paragraph{Complexity of Implementation} 

Assume that the number of instances is \( N \), the number of clusters is \( M \), the number of subcenters in secondary clustering is \( K \), and the dimensionality of features is \( D \). The spatial complexity of the BRST module is \( O(N+M) \) and its temporal complexity is \( O(NMD) \). For MCCL, the spatial complexity is \( O(NMK) \), while the temporal complexity is \( O(NMKD) \). The model contains 98.61 million parameters and requires 32.78 GFLOPs per image for inference.

\paragraph{Visualization} 

To demonstrate the effectiveness of our method in learning modality-invariant features, we randomly select 9 identities from the SYSU-MM01 dataset and visualize their feature embeddings using t-SNE~\cite{van2008visualizing}. As shown in Figure~\ref{fig:4}, our approach generates more compact feature distributions for the same identities within each modality compared to the baseline. This indicates that our method effectively reduces intra-modality variations. Furthermore, the feature embeddings of the same identities across different modalities are significantly closer in proximity, demonstrating improved alignment between visible and infrared features.

To further evaluate the performance of our method, we analyze the distributions of Euclidean distances between randomly selected positive and negative visible-infrared pairs. As shown in Figure~\ref{fig:5}, the variations in $\delta$ demonstrate the effectiveness of our approach in improving cross-modal separability. Specifically, positive cross-modal pairs exhibit a higher degree of convergence, as indicated by their smaller distances, while negative pairs show clear divergence, with larger distances. This suggests that our method with proposed modules effectively enhances the separability between positive and negative pairs in the cross-modal setting.

In addition, we visualize some of the ranking lists on the SYSU-MM01 dataset, as shown in Figure~\ref{fig:8}, and compare the performance of our method with the baseline. The results further demonstrate the effectiveness of our approach in generating high-quality cross-modality pseudo-labels, leading to more accurate ranking outcomes.

\section{Conclusion}
We propose a novel Hierarchical Identity Learning (HIL) framework for visible-infrared person re-identification tasks. HIL leverages Multi-Center Contrastive Learning (MCCL) to enhance intra-modal clustering and minimize cross-modal discrepancies by refining representations through contrastive learning. Additionally, the Bidirectional Reverse Selection Transmission (BRST) mechanism improves cross-modal matching by performing bidirectional pseudo-label matching and filtering unreliable pseudo-label matches. Extensive experiments demonstrate that our approach achieves superior performance in various settings. Our future work could focus on adaptive thresholding and dynamic refinement of pseudo-labels within the BRST mechanism, which could be investigated to improve the accuracy of label assignments.

\bibliographystyle{IEEEtran}
\bibliography{IEEEtran}

\begin{IEEEbiography}[{\includegraphics[width=1in,height=1.25in,clip,keepaspectratio]{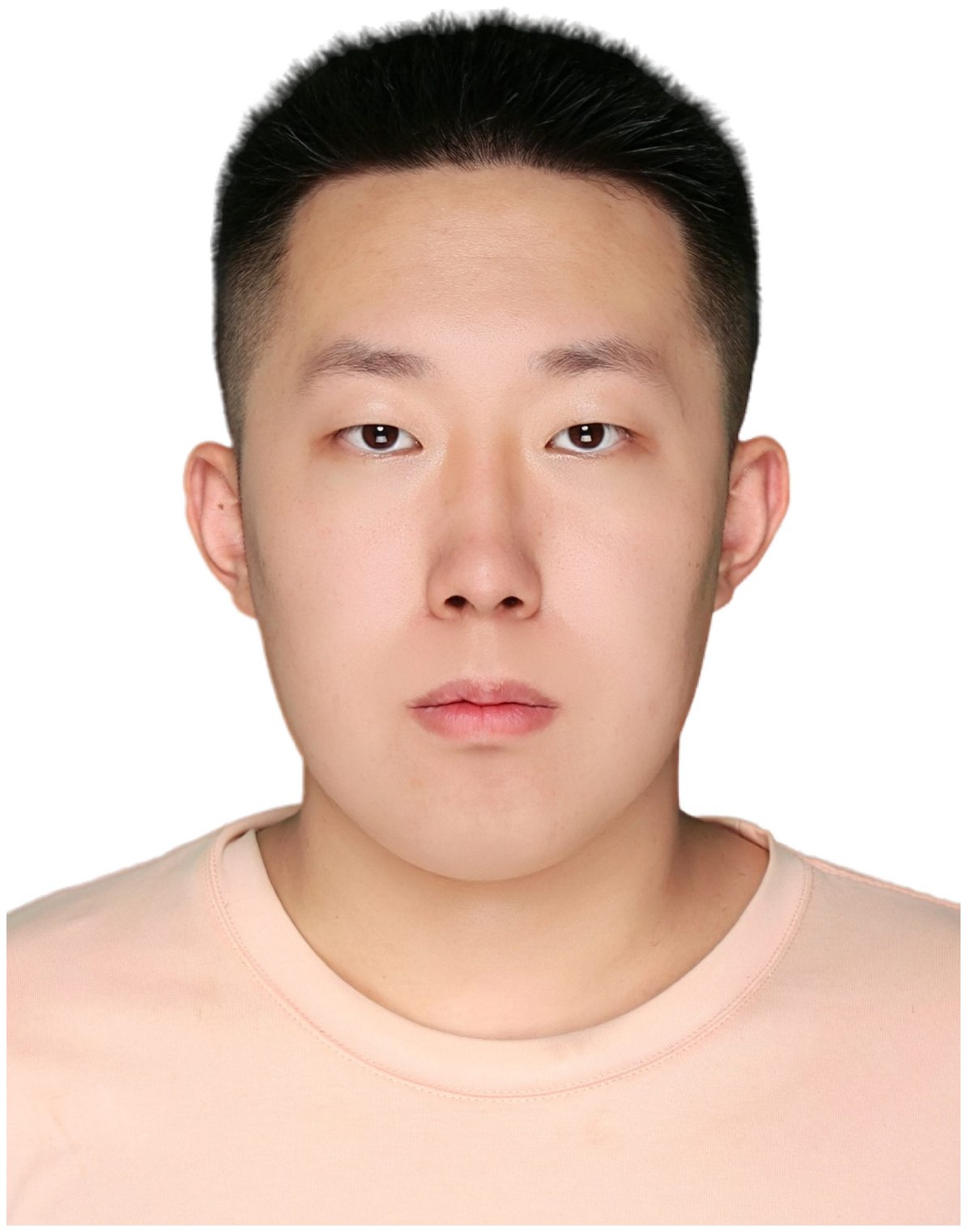}}]{Haonan Shi}
is currently pursuing a B.Sc. degree in Intelligent Science and Technology at Xidian University, Xi'an, China. His research interests include computer vision, pattern recognition, and machine learning.
\end{IEEEbiography}

\begin{IEEEbiography}[{\includegraphics[width=1in,height=1.25in,clip,keepaspectratio]{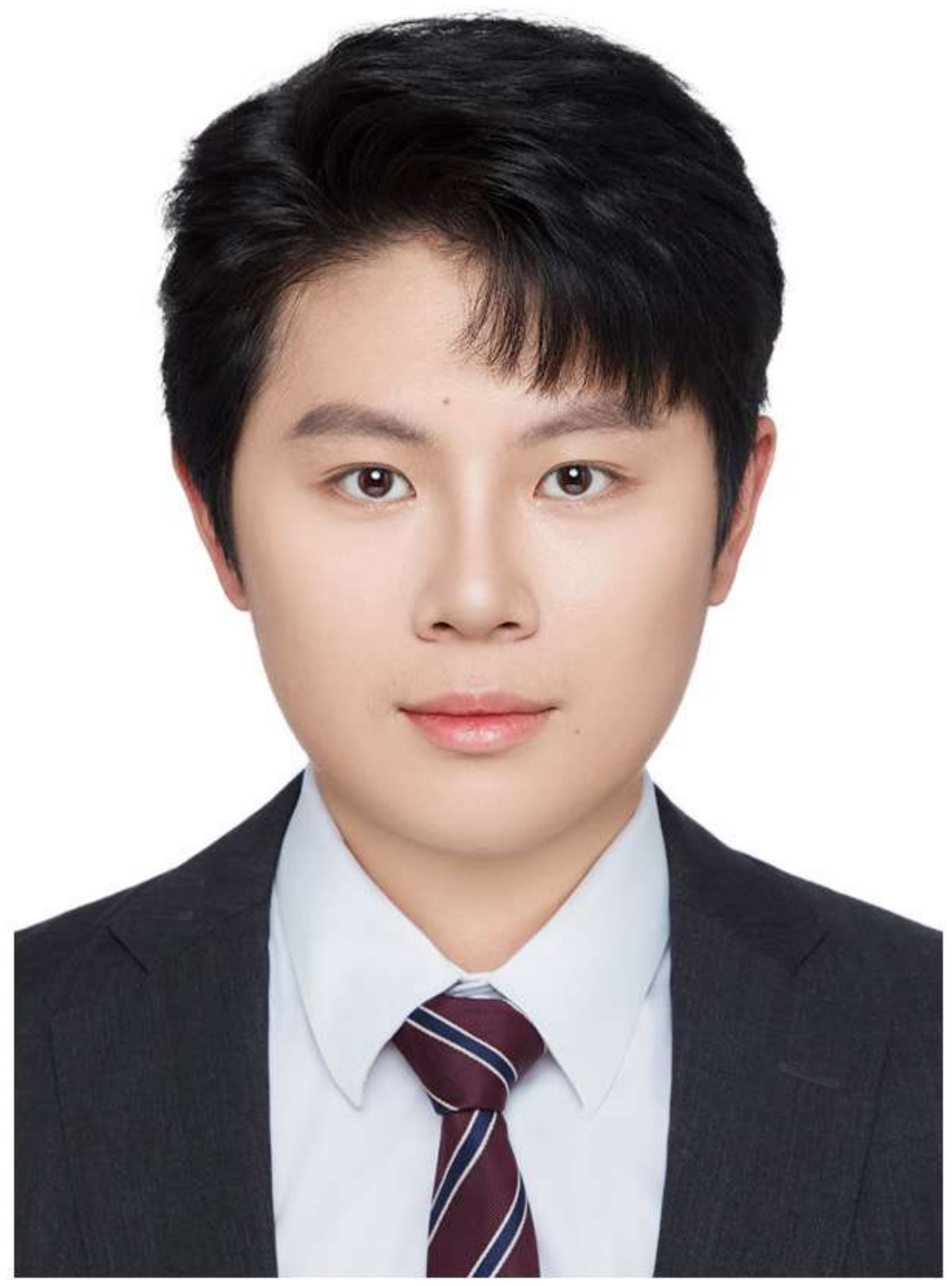}}]{Yubin Wang}
received the B.E. degree in data science and big data technology from Tongji University, China, in 2022, where he is currently pursuing the master’s degree. His main research interests include prompt learning, multi-modal learning, and person re-identification.
\end{IEEEbiography}

\begin{IEEEbiography}[{\includegraphics[width=1in,height=1.25in,clip,keepaspectratio]{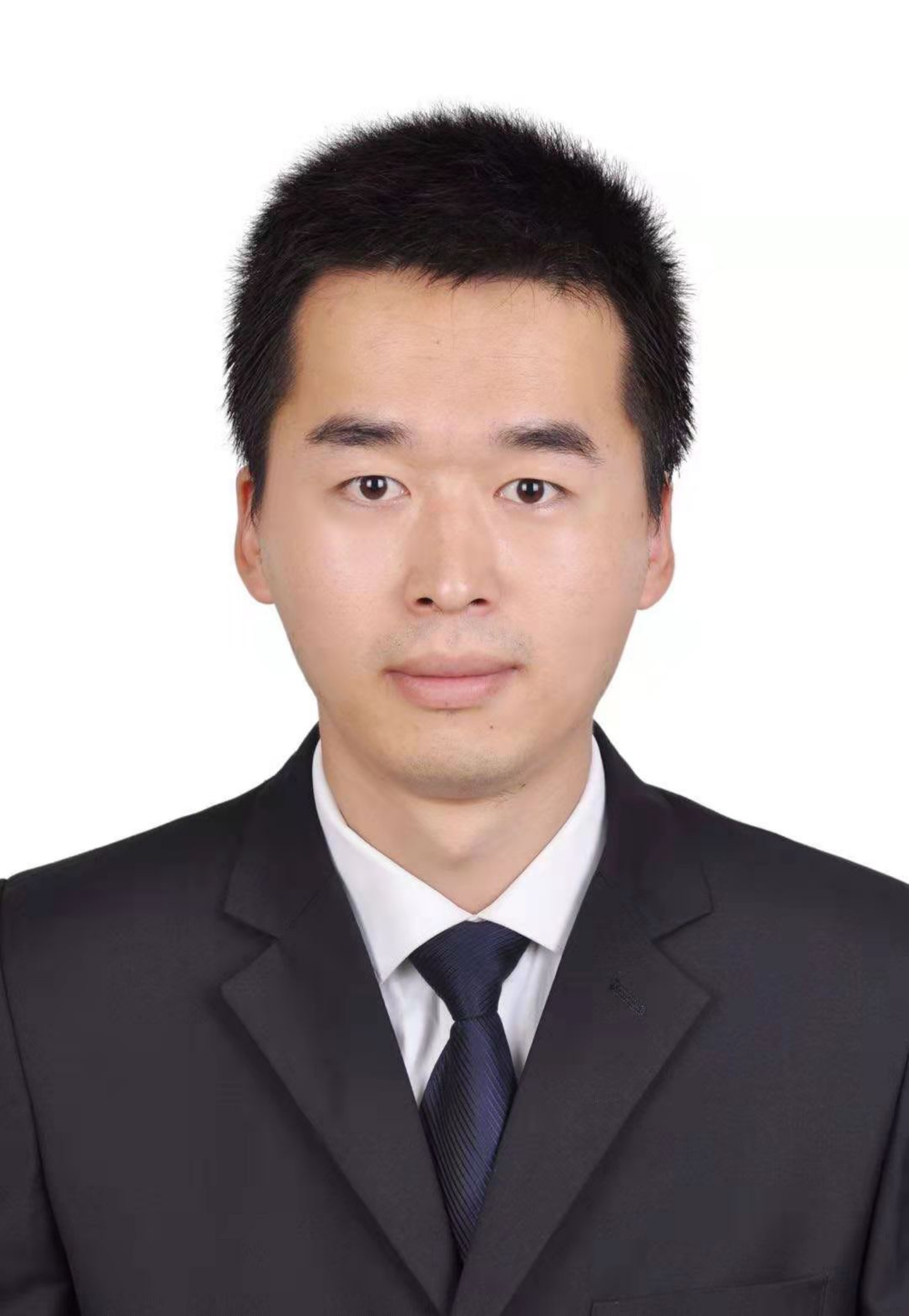}}]{De Cheng} is an associate professor with School of Telecommunications Engineering, Xidian University, China. He received the B.S. and Ph.D. degrees from Xi'an Jiaotong University, Xi'an, China, in 2011 and 2017, respectively. From 2015 to 2017, he was a visiting scholar in Carnegie Mellon University, Pittsburgh, USA. His research interests include pattern recognition, machine learning, and multimedia analysis.
\end{IEEEbiography}



\begin{IEEEbiography}[{\includegraphics[width=1in,height=1.25in,clip,keepaspectratio]{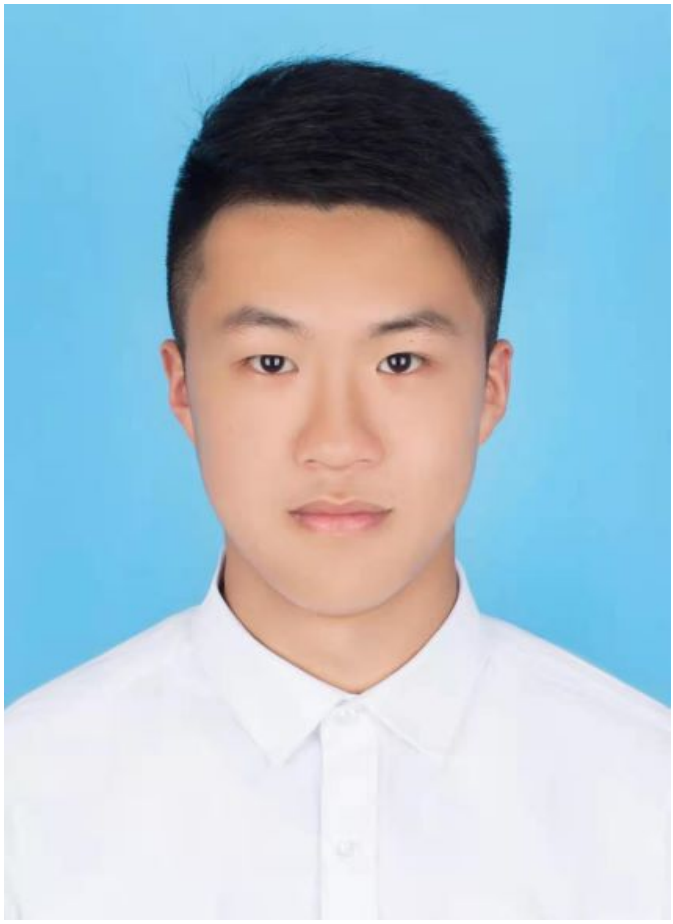}}]{Lingfeng He} received the B.Sc. degree from Xidian University, Xi'an, China, in 2023. He is currently pursuing his M.S. degree in Information and Communication Engineering in Xidian University. His research interests in person re-identification and unsupervised learning.
\end{IEEEbiography}

\begin{IEEEbiography}[{\includegraphics[width=1in,height=1.25in,clip,keepaspectratio]{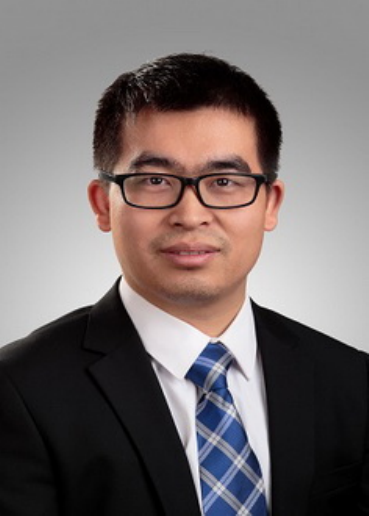}}]{Nannan Wang}
(M'16) received the B.Sc. degree in information and computation science from the Xi'an University of Posts and Telecommunications in 2009 and the Ph.D. degree in information and telecommunications engineering from Xidian University in 2015. From September 2011 to September 2013, he was a Visiting Ph.D. Student with the University of Technology, Sydney, NSW, Australia. He is currently a Professor with the State Key Laboratory of Integrated Services Networks, Xidian University. He has published over 100  articles in refereed journals and proceedings, including IEEE T-PAMI, IJCV, CVPR, ICCV etc. His current research interests include computer vision and machine learning.
\end{IEEEbiography}

\begin{IEEEbiography}[{\includegraphics[width=1in,height=1.25in,clip,keepaspectratio]{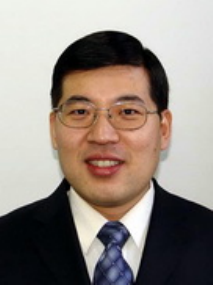}}]{Xinbo Gao}(M'02-SM'07) received the B.Eng., M.Sc. and Ph.D. degrees in electronic engineering, signal and information processing from Xidian University, Xi'an, China, in 1994, 1997, and 1999, respectively. From 1997 to 1998, he was a Research Fellow with the Department of Computer Science, Shizuoka University, Shizuoka, Japan. From 2000 to 2001, he was a Post-Doctoral Research Fellow with the Department of Information Engineering, the Chinese University of Hong Kong, Hong Kong. Since 2001, he has been with the School of Electronic Engineering, Xidian University. He is also a Cheung Kong Professor of the Ministry of Education of China, a Professor of Pattern Recognition and Intelligent System with Xidian University, and a Professor of Computer Science and Technology with the Chongqing University of Posts and Telecommunications, Chongqing, China. He has published 6 books and around 300 technical articles in refereed journals and proceedings. His research interests include image processing, computer vision, multimedia analysis, machine learning, and pattern recognition. Prof. Gao is also a Fellow of the Institute of Engineering and Technology and the Chinese Institute of Electronics. He has served as the general chair/cochair, the program committee chair/co-chair, or a PC member for around 30 major international conferences. He is also on the Editorial Boards of several journals, including Signal Processing (Elsevier) and Neurocomputing (Elsevier).

\end{IEEEbiography}

\end{document}